\documentclass[lettersize,journal]{IEEEtran}
\usepackage{amsmath,amsfonts}

\usepackage{algorithm}
\usepackage{algorithmicx}
\usepackage{algpseudocode} 

\usepackage{array}
\usepackage{textcomp}
\usepackage{stfloats}
\usepackage{url}
\usepackage{verbatim}
\usepackage{graphicx}
\usepackage{cite}
\usepackage[T1]{fontenc}

\usepackage{booktabs}
\usepackage{multirow}

\usepackage{CJK}
\usepackage[top=2cm, bottom=2cm, left=2cm, right=2cm]{geometry}
\usepackage{algorithmicx}
\usepackage{algpseudocode}

\floatname{algorithm 1}{算法}

\usepackage{hyperref}

\hypersetup{hidelinks,
	colorlinks=true,
	allcolors=black,
	pdfstartview=Fit,
	breaklinks=true}

\usepackage{amssymb}
\usepackage{amsthm}

\DeclareMathOperator*{\argmin}{arg\,min}

\makeatletter
\newcommand{\removelatexerror}{\let\@latex@error\@gobble}
\makeatother

\usepackage{bm}

\usepackage{color}
\usepackage{footnote}
\usepackage{threeparttable}

\usepackage{xcolor}

\usepackage{enumerate}

\usepackage{tabularray}
\usepackage{subfigure}

\begin{document}

\title{SRL-ORCA: A Socially Aware Multi-Agent Mapless Navigation Algorithm In Complex Dynamic Scenes}

\author{Jianmin~Qin,
        Jiahu~Qin\textsuperscript{*}, 
        Jiaxin~Qiu,
        Qingchen~Liu,
        Man~Li,    
        and~Qichao~Ma
\thanks{\textsuperscript{*}Corresponding author.}
\thanks{Jianmin~Qin, Jiahu~Qin, Jiaxin~Qiu, Qingchen~Liu, Man~Li, and~Qichao~Ma, are with the Department of Automation, University of Science and Technology of China, Hefei 230027, China (e-mail: qjm@mail.ustc.edu.cn; jhqin@ustc.edu.cn; jiaxinqiu@mail.ustc.edu.cn; qingchen\_liu@ustc.edu.cn; lm1994@mail.ustc.edu.cn; mqc0214@ustc.edu.cn;).
}}

\markboth{Journal of \LaTeX\ Class Files,~Vol.~14, No.~8, August~2022}%
{Shell \MakeLowercase{\textit{et al.}}: A Sample Article Using IEEEtran.cls for IEEE Journals}


\maketitle

\begin{abstract}
For real-world navigation, it is important to endow robots with the capabilities to navigate safely and efficiently in a complex environment with both dynamic and non-convex static obstacles. However, achieving path-finding in non-convex complex environments without maps as well as enabling multiple robots to follow social rules for obstacle avoidance remains challenging problems. In this letter, we propose a socially aware robot mapless navigation algorithm, namely Safe Reinforcement Learning-Optimal Reciprocal Collision Avoidance (SRL-ORCA). This is a multi-agent safe reinforcement learning algorithm by using ORCA as an external knowledge to provide a safety guarantee. This algorithm further introduces traffic norms of human society to improve social comfort and achieve cooperative avoidance by following human social customs. The result of experiments shows that SRL-ORCA learns strategies to obey specific traffic rules. Compared to DRL, SRL-ORCA shows a significant improvement in navigation success rate in different complex scenarios mixed with the application of the same training network. SRL-ORCA is able to cope with non-convex obstacle environments without falling into local minimal regions and has a 14.1\% improvement in path quality (i.e., the average time to target) compared to ORCA. Videos are available at https://youtu.be/huhXfCDkGws.
\end{abstract}


\section{Introduction}

\IEEEPARstart{M}{apless} navigation is an important task for various practical applications in industry, transportation, and public services, such as smart warehousing and autonomous driving. One essential problem of mapless navigation is designing safe and efficient navigation algorithms with limited observations of the dynamic environment. In recent years, the development of deep reinforcement learning (DRL) has achieved remarkable results in a variety of fields \cite{silver2017mastering}. In the field of mapless navigation, DRL is used to train an end-to-end navigation planner for the robot\cite{chen2017decentralized,fan2020distributed}. Due to the dimensional explosion of the action space for multi-agent systems, achieving safe navigation in complex environments with dynamic obstacles remains challenging. In scenarios with high robot density,
the behavior of collision avoidance requires very precise and robust actions of the controller. However, the neural network trained by DRL has poor generalization ability to achieve sufficiently precise collision avoidance, and thus it is difficult to achieve a high navigation success rate\cite{chen2017decentralized}.
Therefore, the study of safe reinforcement learning\cite{garcia2015comprehensive, maclin1996creating} is critical to guarantee the safety of collision avoidance in DRL navigation.

On the other hand, we note that the recent advances in robotics technology are gradually enabling robots to work in shared spaces with humans, such as hospitals, airports, and shopping malls. Therefore, understanding and complying with social rules to cooperate in avoiding collisions and maintaining a safe and comfortable social distance from each other is also an important aspect when designing navigation algorithms \cite{chen2017socially}. Integrating social rules and cooperation in multi-agent navigation can reduce trajectory oscillations or congestion, so as to obtain a relatively optimal global trajectory.

In this work, we propose a socially aware mapless navigation algorithm with SRL-ORCA. It integrates the advantages of RL and conventional collaborative collision avoidance method while considering the social rules. Compared with the existing works, the main contributions and novelties of this article are summarised as follows: 1) In the specific problem of navigation, the SRL-ORCA algorithm implements multi-agent safe reinforcement learning by using ORCA as an external knowledge to provide safety guarantee. Our algorithm significantly improves the success rate of dynamic obstacle avoidance and demonstrates better performance in motion safety of robots over state-of-art work; 2) Our algorithm introduces traffic norms of human society to enhance social comfort and achieve cooperative avoidance that obey human social rules (such as passing on the right, overtaking on the left and low-priority vehicles avoiding high-priority ones). To the best of the authors' knowledge, this is the first work that social rules are considered in such a complicated navigation scenario; 3) SRL-ORCA overcomes the problem that ORCA cannot handle non-convex static obstacles and significantly improves navigation success rate in complex scenes mixed with dynamic and non-convex static obstacles.

\section{RELATED WORK}

For mapless navigation tasks in complex dynamic environments, three elementary tasks need to be considered: finding feasible paths to destinations (Task-1); obeying common social rules (e.g., traffic rules) among the agents (Task-2); and achieving dynamic and static obstacle avoidance (Task-3). Existing studies can be broadly classified into three categories: learning-based navigation methods, conventional navigation methods, and their fusion methods.

\subsection{Learning-based navigation approaches}

Recently, learning-based navigation demonstrates better performance and adaptability than conventional methods. In \cite{pfeiffer2017perception}, a target-oriented motion planning module is trained based on expert demonstrations. However, the performance of the learning-based algorithms is severely limited by the quality of the labeled training set. To overcome this deficiency, \cite{chen2017decentralized} proposes a decentralized multi-agent collision avoidance algorithm, which shows more than $26\% $ improvement in path quality compared to ORCA. To guarantee safe exploration during training, Yan et al.\cite{yan2022mapless} introduce a novel imitation learning training scheme based on Dataset Aggregation to obtain a mapless navigation policy.

\subsection{Conventional navigation approaches}

Conventional navigation algorithms are usually a combination of global planning (e.g., A*, D* \cite{dolgov2008practical}) and local collision avoidance (e.g., DWA\cite{fox1997dynamic}, VO\cite{fiorini1998motion}, ORCA\cite{berg2011reciprocal}). The major challenge of mapless navigation is to achieve collision avoidance safely in scenes mixed with dynamic and static obstacles. However, global planning algorithms usually do not work accurately without maps, which can easily cause navigation strategies to fall into local minimal regions and fail\cite{foead2021systematic}. In recent years, ORCA\cite{berg2011reciprocal,alonso2013optimal} is widely used in multi-robot systems and crowd simulation. But ORCA has two shortcomings: First,
it is limited to dealing with convex obstacles, and it cannot even bypass simple non-convex static obstacles and easily gets trapped in local minimal areas. Second, in scenarios with high robot density, robots using ORCA lack cooperation in collision avoidance and tend to form congestion or trajectory oscillations, which reduce the average arrival time of all robots\cite{chen2017decentralized,chen2017socially}.

\subsection{Fusion algorithms of RL and conventional navigation methods}

Currently, there are also some works combining RL and conventional navigation methods\cite{patel2021dwa,zhou2022navigating,xie2023drl,han2022reinforcement}, but none of them can absolutely guarantee dynamic collision avoidance safety. DWA-RL\cite{patel2021dwa} combines the advantages of RL and DWA and achieves better results than RL navigation. Nevertheless, DWA\cite{fox1997dynamic} cannot absolutely guarantee dynamic collision avoidance safety, and the success rate of DWA-RL in scenarios with high-density dynamic obstacles is drastically reduced ($40\%$). HGAT-DRL\cite{zhou2022navigating} uses VO\cite{fiorini1998motion} to describe the observed information of obstacles, making it easier to learn collision avoidance and get a better success rate. DRL-VO\cite{xie2023drl} proposes a novel DRL navigation in pedestrian-filled environments, where VO is used to design the reward function. Han et al. \cite{han2022reinforcement} proposed RL-RVO, which uses the concept of RVO\cite{van2008reciprocal} to design the reward function. The experimental scenario of RL-RVO is rather simple, it is not tested in scenes with a mixture of dynamic and static obstacles. HGAT-DRL, DRL-VO, and RL-RVO all use neural networks as navigation controllers, which cannot absolutely guarantee sufficient generalization performance in dense dynamic obstacle scenes.
Due to the safety shortcomings of neural networks in RL, we need to design a safe reinforcement learning method. It should absolutely guarantee that no catastrophic collisions occur for navigation in complex dynamic environments.

\subsection{Inducing Social Norms in Navigation}

Humans tend to cooperate and adopt traffic rules to avoid collisions (Task-2). This motivates the use of social norms that enable robots to interact with humans in a socially compliant way. 
In conventional approaches, the work in \cite{ferrer2013robot} introduces additional parameters to interpret social interactions and establish social comfort criteria. But they have been observed to potentially lead to oscillatory paths \cite{chen2017decentralized}, \cite{kretzschmar2016socially}. In contrast, learning-based methods recover the expert policy directly through supervised learning \cite{kim2016socially}, or aim to use RL or inverse reinforcement learning to model the complex interactions and cooperation among agents \cite{chen2017socially}, \cite{choi2020fast}. For example, Chen et al.\cite{chen2017socially} designed a complex negative reward to train a time-efficient navigation policy that accords with common social norms.

\section{PROBLEM FORMULATION}

In this section, we formulate multi-robot non-communicating mapless navigation as a partially observable sequential decision problem. In SRL-ORCA, both the DRL and ORCA parts can be uniformly specified as a Markov decision process (MDP). This makes the theoretical foundation for integrating these two methods into our algorithm. In the MDP process, we define a 5-tuple $(S,A,P,R,\gamma)$, where $\gamma\in \left ( 0,1 \right ) $ is the discount factor. There are a total of $M $ moving agents in the scene. For the $i$-th agent $\left ( 1\leq i\leq M \right )$, there are $k $ agents within its observable range that are equivalent to dynamic obstacles.

\textbf{State space:} Similar to \cite{chen2017decentralized}, the state vector of agent-$i$ can be divided into three parts at each timestep $t$, it is $s_{i,t}= [\,  s_{i,t}^{in}, ~\tilde{s}_{i,t}, ~s_{i,t}^{env} \, ]$, where $s_{i,t}^{in} $ denotes the agent's internal state information, $\tilde{s}_{i,t}= [ \, \tilde{s}_{i,t}^{1},~\tilde{s}_{i,t}^{2},\cdots ,~\tilde{s}_{i,t}^{k}\,  ] $ denotes the state of other $k $ dynamic obstacles observed by the agent $i$, and $s_{i,t}^{env} $ denotes the data of static obstacles in the vicinity measured through sensors (e.g., LIDAR). For the agent $i$, we let $d_{i,g} $ represent the destination; let the position and velocity vectors be denoted by $p_{i}$ and $v_{i}$, respectively; let $v_{i,pref} $ be agent’s preferred speed \cite{berg2011reciprocal}; let $\psi_{i} $ represent the directional angle of moving forward; let $r_{i,safe} $ represent the safety radius; let $pr_{i} $ be the driving priority level, and low priority agents should actively avoid high priority ones. Therefore, the internal state is $s_{i,t}^{in}= [\, d_{i,g},~p_{i},~v_{i},~v_{i,pref},~\psi_{i},~r_{i,safe},~pr_{i} \,  ]$. The state of $k $-th dynamic obstacle known by agent $i$ is the position, speed, direction, safety radius and priority level, $\tilde{s}_{i,t}^{k}= [ \,  \tilde{p}_{i}^{k},~\tilde{v}_{i}^{k},~\tilde{r}_{i,safe}^{k},~\tilde{\psi}_{i}^{k},~\tilde{pr}_{i}^{k} \, ]$,  while $\tilde{d}_{i,g}^{k} $ and $\tilde{v}_{i,pref}^{k} $ are hidden states which cannot be observed.

\textbf{Action space:} We implement the method on a  differential driving robot. As in \cite{alonso2013optimal}, the robot's action space should be within the scope of the set $S_{AHV_{i}}^{RL} $(shown in Fig.~3(a)). The action of the robot $i$ is a set of allowed velocities, including translational velocity $ v_{i,t} $ and rotational velocity $ w_{i,t} $, i.e., $a_{i,t}= [\,  v_{i,t} ,w_{i,t} \, ],\: a_{i,t}\in S_{AHV_{i}}^{RL} $. The differential robot can perform continuous actions directly but is not allowed to move backward $ \left ( v_{i,t}> 0 \right ) $.

The design of the reward function will be discussed in detail in Section \uppercase\expandafter{\romannumeral4}-A(2). The probabilistic state transfer model $P\left ( s_{i,t+1},~s_{i,t}| a_{i,t} \right ) $ is unknown.

In SRL-ORCA, we define two independent policy policies for all agents: 1) $\pi _{\theta}^{RL} $ in DRL; 2) $\pi ^{orca} $ in ORCA. Two sets of policies $\pi _{\theta}^{RL} $ and $\pi ^{orca} $ run in parallel. For DRL navigation, the $i$-th agent has access to a state  $s_{i,t}$ and independently computes an action  $a_{i,t}^{RL} $, sampled from a random policy $\pi _{\theta }^{RL}$ shared by all agents at time step $t$: 
\begin{small}
\begin{equation} \mathbf{a}_{i,t}^{RL}\sim\pi _{\theta}^{RL}(\mathbf{a}_{i,t}\vert \mathbf{s}_{i,t}),  
\end{equation} 
\end{small}

where $\theta $ is the policy parameter. Action $a_{i,t} $ drives the agent to safely approach the destination $d_{i,g}$ from the current position $p_{i,t}$ while avoiding collisions with other agents and the obstacle $O_{x}\left ( 0\leq x\leq N \right )$. For ORCA navigation, based on state $s_{i,t} $, agents' priorities $pr_{i} $ and $pr_{j} $, a priority allowed speed set $ORCA_{i,t}^{\tau -prior} $ can be computed for the $i $-th agent (as in Section \uppercase\expandafter{\romannumeral4}-B(1)).

In the fusion algorithm, we use ORCA as an external knowledge to provide safety advice ($ORCA_{i,t}^{\tau -prior} $) to DRL and guarantee safety. 
On the basis of action $a_{i,t}^{RL} $ and safety advice $ORCA_{i,t}^{\tau -prior} $,
We design a switching method to get their fusion policy $\pi _{\theta}^{srl-orca} $ according to the scenario and avoid unsafe actions. In particular, the fused policy $\pi _{\theta}^{srl-orca} $ is expected to follow social rules. We will consider several common human \textbf{social traffic norms:} \textbf{1. overtake on the left.}\textbf{ 2. pass on the right.} \textbf{ 3. low priority vehicles avoid high priority ones }(e.g., Fire engines and ambulances have priority over other traffic). Afterward, we use \textbf{Norm-1,2,3 } to denote the three types of social norms respectively.

\begin{figure}[!t]
	\centering
   	\vspace{-0.0cm}	\includegraphics[width=3.3in]{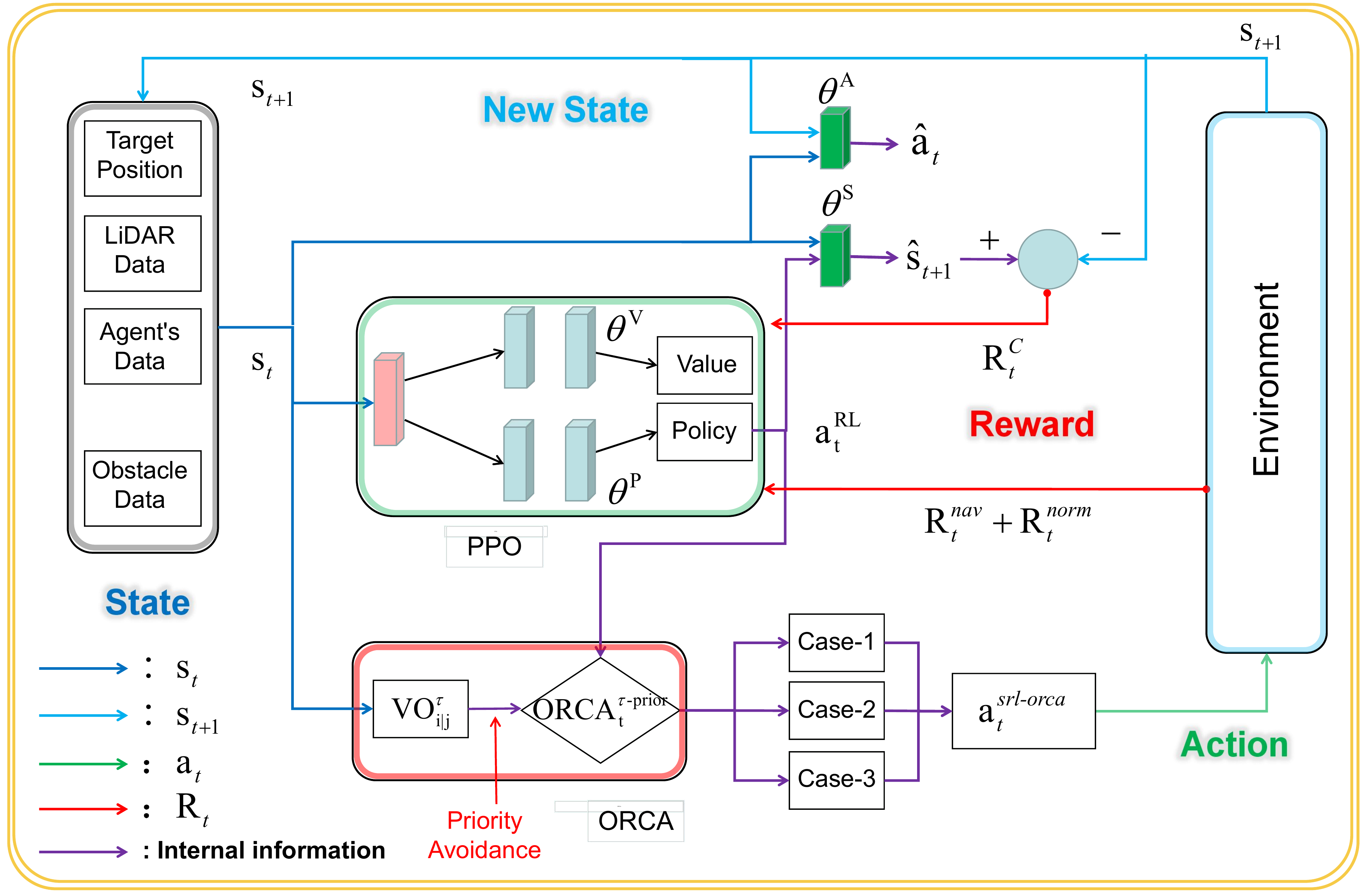}
   	\vspace{-0.0cm}
	\caption{In SRL-ORCA, two sets of algorithms, Priority-ORCA and DRL, operate in parallel. According to state information $s_{i,t} $, DRL uses common policy $\pi_{\theta }^{RL} $ to output action $a_{i,t}^{RL} $ that follows social norms for agent-$i$, and Priority-ORCA calculates the priority permitted speed set $ORCA_{i,t}^{\tau -prior} $. Based on $a_{i,t}^{RL} $ and $ORCA_{i,t}^{\tau -prior} $, we classify and calculate the safe ultimate action $a_{i,t}^{srl-orca} $. For notation simplicity and clear presentation, we omit superscript $i$
in Fig.~1.}
  	\vspace{-0.5cm}
	\label{fig3}
\end{figure}

\section{APPROACH}

We start by introducing the overall structure of our method, as shown in Fig.~1. According to state information $s_{i,t} $, DRL uses policy $\pi_{\theta }^{RL} $ to output action $a_{i,t}^{RL} $ that follows social norms for agent-$i$, and ORCA calculates the priority permitted speed set $ORCA_{i,t}^{\tau -prior} $ (safety advice). Based on $a_{i,t}^{RL} $ and $ORCA_{i,t}^{\tau -prior} $, we classify and calculate the safe ultimate action $a_{i,t}^{srl-orca} $ (as in Section \uppercase\expandafter{\romannumeral4}-B(2)). If action ${a}_{i,t}^{RL} $ is within the scope of safety advice from external knowledge (ORCA), the action ${a}_{i,t}^{RL} $ of DRL  is safe, and we can let ${a}_{i,t}^{srl-orca}={a}_{i,t}^{RL} $. When unsafe situations are encountered, the action ${a}_{i,t}^{RL} $ is modified based on the advice of external knowledge (ORCA) to obtain action ${a}_{i,t}^{srl-orca} $ for avoiding catastrophic collisions. In complex environments mixed with dynamic and static obstacles, this will improve trajectory quality (e.g., average arrival time) and better cope with complex collision avoidance.

For the DRL part, it is difficult to design a reward function to satisfy all situations in countless real-world navigation environments (e.g., mazes and corridors). Sometimes the external rewards for agents are very sparse. Thus, DRL methods are often unable to explore the environment efficiently or obtain better navigation strategies. To handle the problem of sparse rewards, we use a curiosity-driven approach based on self-supervised prediction to enhance agent exploration during training\cite{pathak2017curiosity}. The prediction error in the learning feature space (the difference between the predicted states and true states, i.e., curiosity) is used as an intrinsic reward signal that enables the agent to actively explore the surroundings and learn policy $\pi _{\theta}^{RL} $ that may be effective in subsequent navigation.
We adopt a centralized training (as in \textbf{Algorithm 1}) and distributed execution paradigm (as in \textbf{Algorithm 2}). Next, SRL-ORCA is presented in two parts.

\subsection{Multi-Agent Deep Reinforcement Learning}

In the DRL-part, we use multi-agent DRL to train our navigation policy to produce an action $a_{i,t}^{RL} $ that follows the social norms for agent-$i$. Compared to conventional planning algorithms  (e.g., A* or D*\cite{dolgov2008practical}) that require global maps, DRL can navigate in complex scenes without global maps at low sensor levels (e.g., sparse LiDAR only). The policy $\pi _{\theta}^{RL }:\left ( s_{i,t} \right )\rightarrow a_{i,t}^{RL} $ obtained by training performs more robustly and effectively in environments mixed with dynamic and non-convex static obstacles, and it is more "intelligent" to cope with maze-like scenarios \cite{mirowski2016learning}.

\textbf{(1) Neural network architecture with curiosity module:} 
The network architecture of each agent for SRL-ORCA is shown in Fig.~1, and we adopt PPO-clip \cite{schulman2017proximal} as the DRL training algorithm. We designed two 2-hidden layer neural networks (policy network $\theta ^{P} $ and value network $\theta ^{V} $) as nonlinear function approximators of policy $\pi _{\theta } $ and value function $V _{\theta } $. Each hidden layer is a fully connected layer with 128 rectifier units and the ReLU nonlinearity is applied as the hidden layer activation function. Meanwhile, we used curiosity networks \cite{pathak2017curiosity} to generate internal rewards to strengthen exploration under sparse rewards in complex environments. We apply two fully connected neural networks (action estimation network $\theta ^{A} $ and state estimation network $\theta ^{S} $) for the prediction of action $\hat{a}_{t} $ and state $\hat{s}_{t+1} $, respectively. After the network is constructed, the input is the observation of each robot and the output is the action of the decentralized robots.

The training process of SRL-ORCA is as follows (as in \textbf{Algorithm 1}). All of the $M$ agents share a common network structure (the 4 networks, $\theta ^{P},\theta ^{V},\theta ^{A},\theta ^{S} $). During training, their states $s_{i,t} $ are sequentially input to the networks. In this way, the shared navigation policy $\pi _{\theta }^{RL} $ of $M$ agents can be obtained by learning. Each agent takes action under current policy $\pi _{\theta }^{RL} $ and generates trajectories, then updates the policy $\pi _{\theta }^{RL} $ based on the sampled data of the trajectories.

For network $\theta ^{A} $, the input state $s_{i,t} $ is encoded as feature vector $\lambda \left ( s_{i,t} \right ) $. Then we predict that the agent takes the action $\hat{a}_{i,t} $ to transfer from state $s_{i,t} $ to $s_{i,t+1} $  based on the input feature vectors $\lambda \left ( s_{i,t} \right ) $ and $\lambda \left ( s_{i,t+1} \right ) $. From the inverse dynamics function $f\left ( \cdot  \right ) $, we can obtain the predicted action $\hat{a}_{i,t}=f\left (s_{i,t},s_{i,t+1}:\theta ^{A} \right ) $. We optimize network $\theta ^{A} $ to reduce the loss function  $L_{A} $ that measures the difference between predicted actions and actual actions:

\begin{small}
  	\vspace{-0.3cm}
	\begin{equation}
		\underset{\theta^{A}}{\min} \: L_{A} \left \{ \left ( \hat{a}_{i,t},a_{i,t}\right )|\hat{a}_{i,t}=f\left ( s_{i,t},s_{i,t+1}:\theta ^{A} \right )  \right \}.
	\end{equation}
   	\vspace{-0.3cm}
\end{small}

For network $\theta^{S} $, we input $a_{i,t} $ and  $\lambda \left (s_{i,t}  \right ) $ to train $\theta^{S} $ to predict the feature vector  $\hat{\lambda} \left (s_{i,t+1}  \right ) $ of the state on time step $t+1 $, which gives the forward dynamics function  $h\left ( \cdot  \right ) $ as:  $\hat{\lambda} \left (s_{i,t+1}  \right )=h\left ( \lambda \left (s_{i,t}  \right ),a_{i,t}:\theta ^{S} \right ) $. The network $\theta^{S} $ is optimized by minimizing the difference between $\hat{\lambda} \left (s_{i,t+1}  \right ) $ and $\lambda \left (s_{i,t+1} \right ) $. The minimization loss function  $L_{S} $ is

\begin{scriptsize}
  	\vspace{-0.2cm}
	\begin{equation}
         \underset{\theta^{S}}{\min} \: L_{S} \left \{ \left ( \lambda \left (s_{i,t+1} \right ),\hat{\lambda} \left (s_{i,t+1}  \right ) \right )|\hat{\lambda} \left (s_{i,t+1}  \right )=h\left ( \lambda \left (s_{i,t}  \right ),a_{i,t}:\theta ^{S} \right ) \right \}.
	\end{equation}    
\end{scriptsize}

When agents explore new environments, the predicted feature vector $\hat{\lambda} \left (s_{i,t+1}  \right ) $ is significantly different from the real $\lambda \left (s_{i,t+1} \right ) $. Exploring new environments gives agents a "curiosity" reward, which is necessary for agents to enhance exploration to seek future rewards. Following this principle, we can design the internal curiosity reward as

\begin{small}
  	\vspace{-0.2cm}
	\begin{equation}
		R_{t}^{C}=\frac{\delta }{2}\left \| \hat{\lambda} \left (s_{i,t+1}  \right )- \lambda \left (s_{i,t+1} \right ) \right \|_{2}^{2},
	\end{equation}
\end{small}

\noindent where $\delta $ is the curiosity strength factor. Curiosity is a new mode of learning, and when extrinsic rewards are sparse, intrinsic rewards are important to motivate agents to explore new environments and discover novel states in order to obtain better learning policies. 
As in \textbf{Algorithm 1}, line 10, the overall optimization problem solved by the agent is the joint optimization of PPO's policy update formulation \cite{schulman2017proximal} and loss functions Eq.(2) and Eq.(3), which can be defined as:

\begin{small}
  	\vspace{-0.4cm}
	\begin{equation}
		\underset{\theta ^{P},\theta ^{A},\theta ^{S}}{\min}\left [ -\alpha \underset{s,a\sim \pi _{\theta_{y}^{P} }}{\mathbb{E}}\left [ L\left ( s,a,\theta_{y}^{P},\theta ^{P} \right ) \right ]+\left ( 1-\beta  \right )L_{A}+\beta L_{S} \right ],
	\end{equation}
\end{small}

\noindent where $\alpha > 0 $ is a scaling factor to evaluate the importance of the strategy gradient, $\theta_{y}^{P} $ is the network of the $y$-th iteration of network $\theta^{P} $, $1\geq \beta  \geq  0 $ is a factor to weight the predicted action loss $L_{A} $ and the predicted state feature vector loss $L_{S} $. The training of the value network $\theta^{V} $ follows the PPO-clip approach and still fits the value function by regression on mean square error \cite{schulman2017proximal}.

\begin{figure}[!htb]
 	\vspace{-0.6cm}
	\removelatexerror
	\begin{algorithm}[H]
		\caption{The training algorithm of SRL-ORCA}
		\begin{algorithmic}[1]
			\Require shared networks $\theta^{P} $, $\theta^{V} $, $\theta^{A} $, $\theta^{S} $. Group of agents $i\in \left [ 1,M \right ] $ provide with: set $S_{AHV_{i}}^{RL} $, priority level $pr_{i} $.
			\For{iteration = 1,2,...,}
			\For{agent $i\in \left [ 1,M \right ] $}
			\State Run policy $\pi _{\theta }^{RL} $ for $T $ timesteps.
			\State Obtain action $a^{RL}_{i}=\vec{v}^{RL}_{i} $ by Algorithm-2.
			
			\State Compute curiosity rewards $R_{t}^{C} $ by Eq. (4).
			
			\State Collect $\left \{ s_{i,t},\,r_{i,t},\,a_{i,t} \right \}$, where $t\in \left [ 0, \,T \right ] $.
			
			\State Compute advantage estimates as in PPO-Clip.
			
			\EndFor
			\State $\pi _{old}^{RL}\leftarrow \pi _{\theta }^{RL} $.
			\State Joint optimal policy $\theta^{P} $, $\theta^{A} $, $\theta^{S} $ as in Eq.(5).
			\State Update $\theta^{V} $ as in PPO-Clip.
			\EndFor
			
		\end{algorithmic}
		\label{alg:Training}
	\end{algorithm}
  	\vspace{-0.5cm}
\end{figure}

\begin{figure*}[htbp]
	\centering
	\vspace{-0.0cm}
\centering
\subfigure[]{
\begin{minipage}[b]{0.18\linewidth}
\centering
\includegraphics[width=1\textwidth]{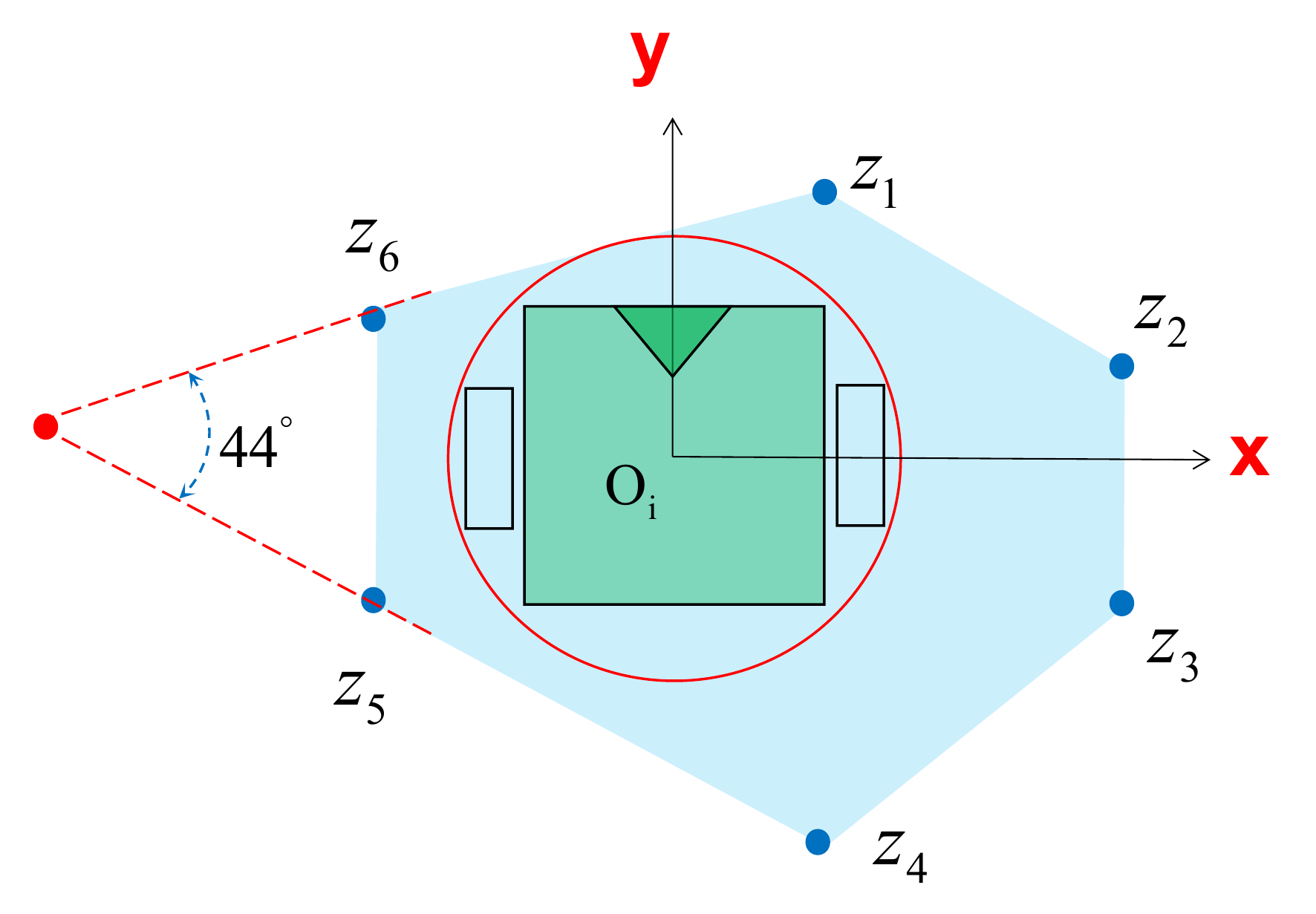}
\end{minipage}%
}%
\subfigure[]{
\begin{minipage}[b]{0.2\linewidth}
\centering
\includegraphics[width=1\textwidth]{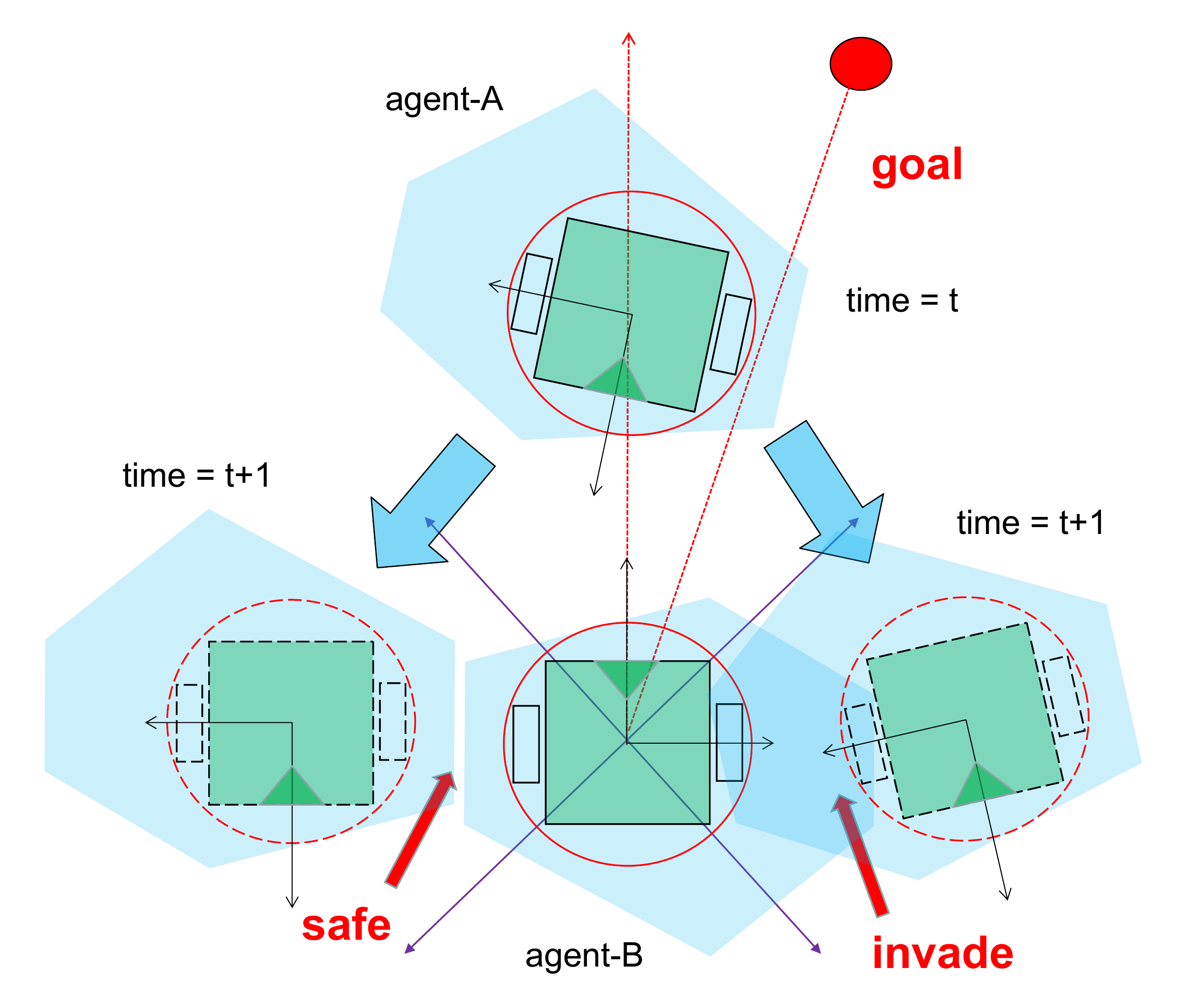}
\end{minipage}%
}%
\subfigure[]{
\begin{minipage}[b]{0.2\linewidth}
\centering
\includegraphics[width=1\textwidth]{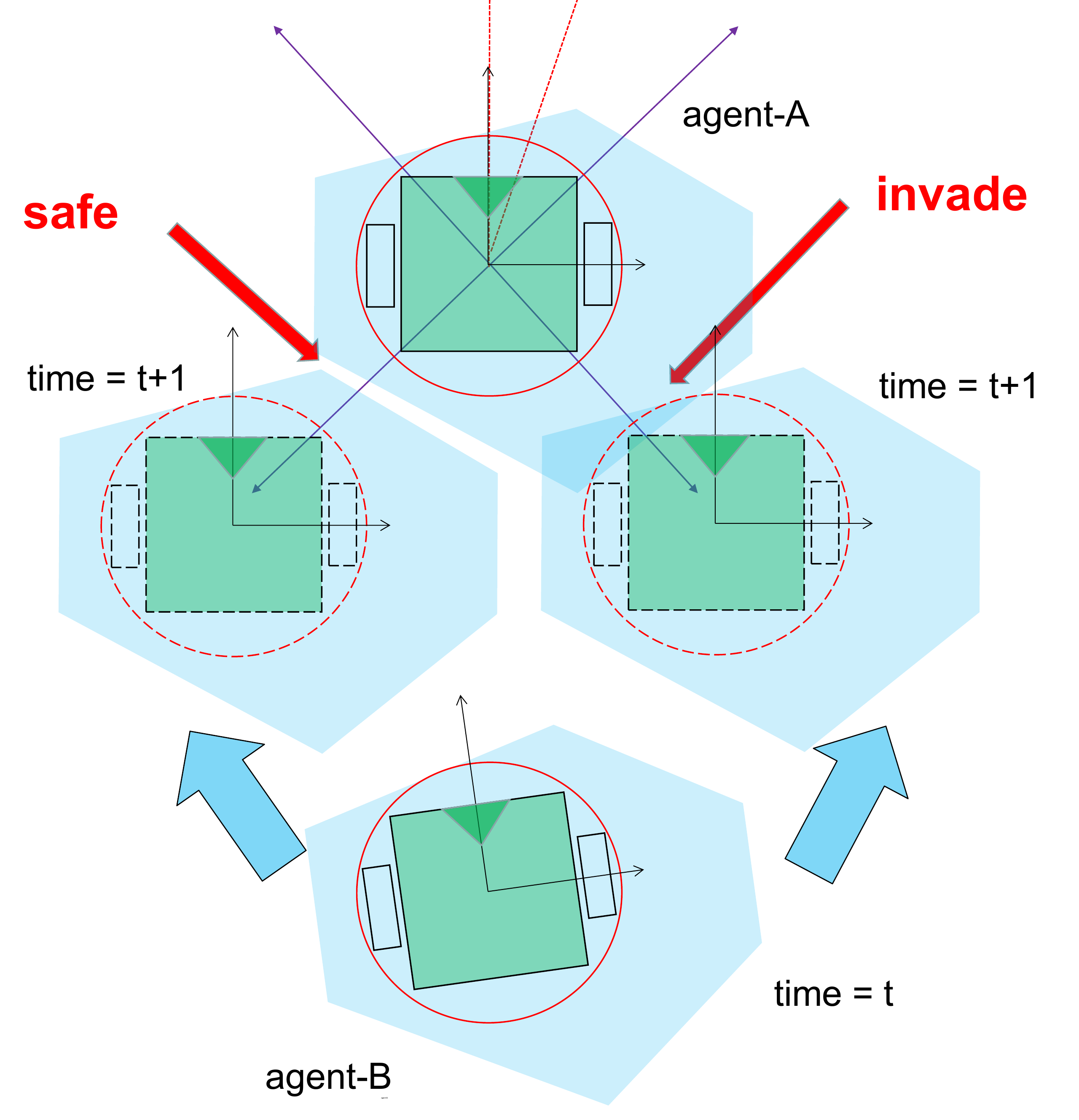}
\end{minipage}%
}%
\subfigure[]{
\begin{minipage}[b]{0.2\linewidth}
\centering
\includegraphics[width=1\textwidth]{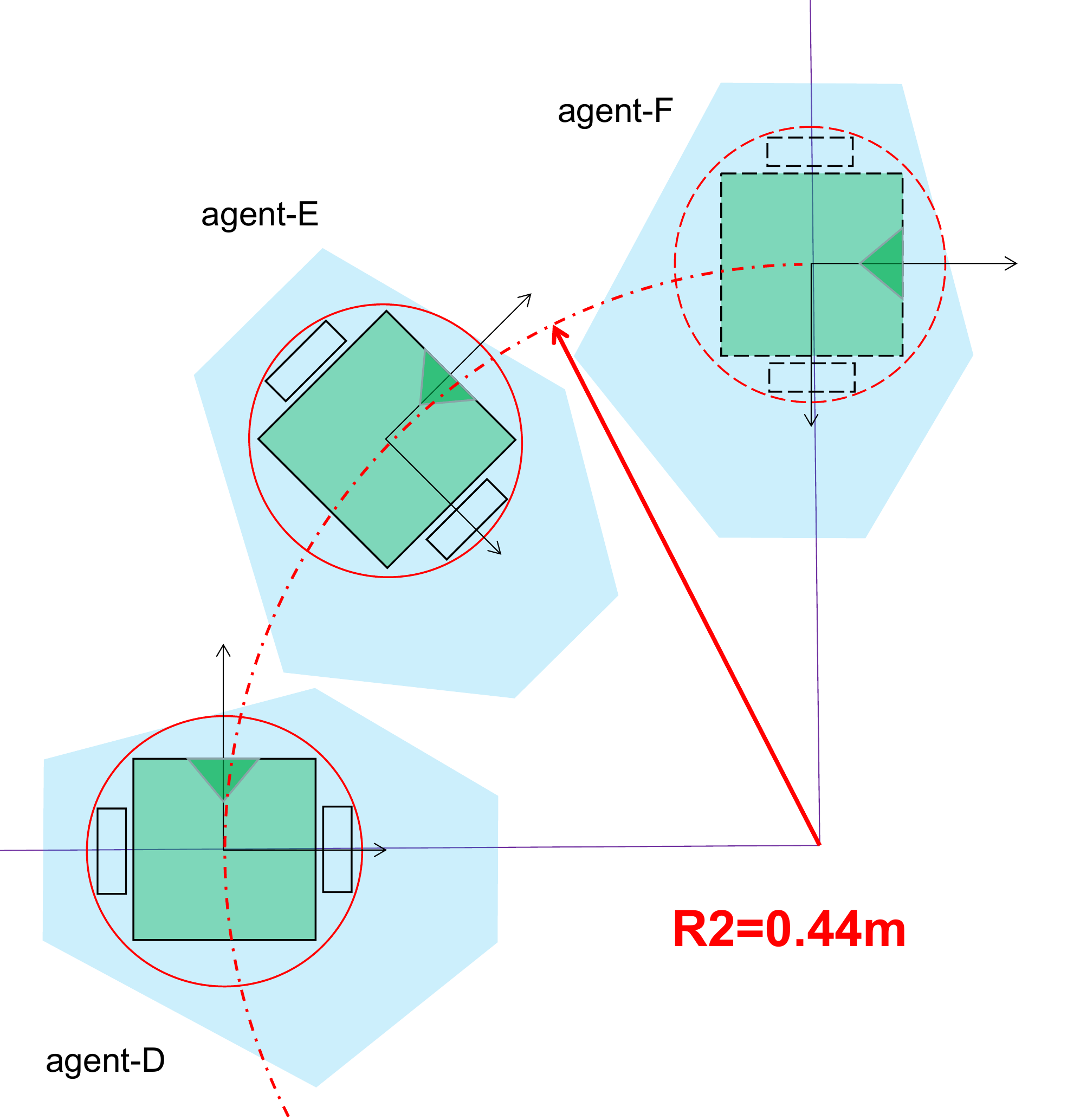}
\end{minipage}
}%
\subfigure[]{
\begin{minipage}[b]{0.2\linewidth}
\centering
\includegraphics[width=1\textwidth]{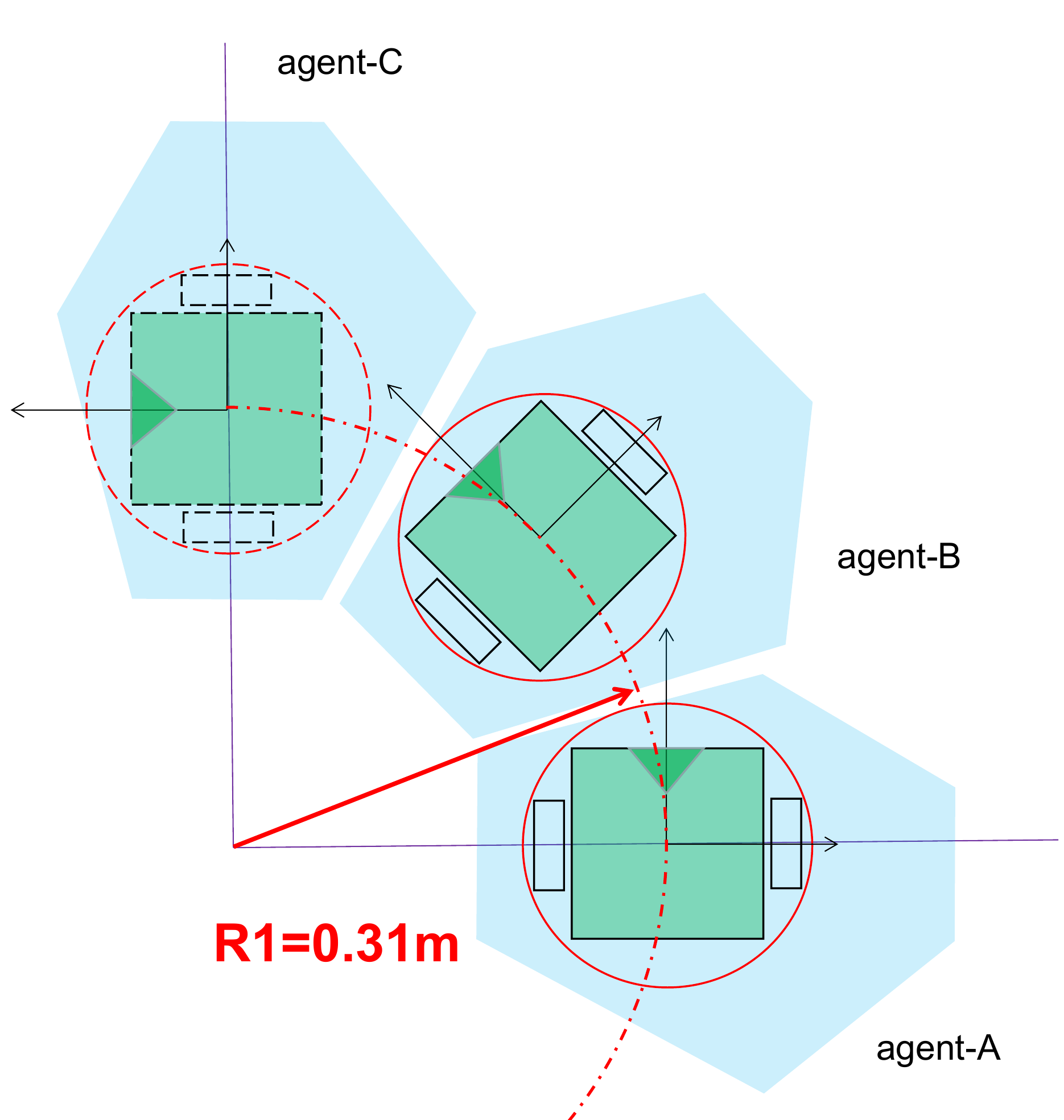}
\end{minipage}
}%
\centering

\label{fig5b}
	\vspace{-0.2cm}
\caption{(a) For a differential wheeled robot-$i$ with safety radius $r_{i,safe}=0.105 \, m $ $\left ( 1\leq i\leq M \right )$, the light blue area is the asymmetric social norm penalty area $area_{i,norm} $, ; (b) is the case of agents meeting head-on; (c) is the case of agents overtaking; (d) is clockwise; (e) is anti-clockwise. When an agent's own penalty area $area_{i,norm} $ overlaps with another agent's penalty area $area_{j,norm} $ ($i\neq j,1\leq i\leq M,1\leq j\leq M $), both of them will receive a slight penalty reward $R_{t}^{norm} $. In our experiments, we choose these key points for $area_{i,norm} $, $z_{1}=\left ( 0.07m,0.126m \right )$, $z_{2}=\left ( 0.21m,0.07m \right )$, $ z_{3}=\left ( 0.21m,-0.07m \right )$, $z_{4}=\left ( 0.07m,-0.182m \right )$, $z_{5}=\left ( -0.14m,-0.07m \right )$, $z_{6}=\left ( -0.14m,0.07m \right ) $.}
 	\vspace{-0.5cm}
	\label{fig5}
\end{figure*}

\textbf{(2) Reward design with social norms:} 
The external reward function $R_{t}^{ex} $ is divided into two parts: navigation reward  $R_{t}^{nav} $ and  norm reward $R_{t}^{norm} $. $R_{t}^{nav} $ is to guide agents to explore the path to the goal without collision and minimize the average arrival time. $R_{t}^{norm} $ induces agents to learn to follow social rules by introducing small biases,

\begin{small}
  	\vspace{-0.2cm}
\begin{equation} 
R_{t}^{ex}=R_{t}^{nav}+R_{t}^{norm}, \end{equation}   
  	\vspace{-0.1cm}
\end{small}
\begin{small}
  	\vspace{-0.2cm}
\begin{equation} R_{t}^{nav}=R_{t}^{mf}+R_{t}^{dir}+R_{t}^{col-s}+R_{t}^{col-d}+R_{t}^{tim}+R_{t}^{goal},  
\end{equation}  
  	\vspace{-0.0cm}
\end{small}
where $R_{t}^{mf} $ is a reward for moving forward;  $R_{t}^{dir} $ represents a reward for the angle between the agent's movement direction and the target's movement direction, with a reward for the correct direction and a penalty for the opposite:

\begin{small}
  	\vspace{-0.2cm}
	\begin{equation}
		R_{t}^{mf}=b\cdot \left ( \left \| d_{i,g}-p_{i,t=0} \right \|  - \left \| d_{i,g}-p_{i,t=t_{now}} \right \|\right ),
	\end{equation}
   	\vspace{-0.2cm}
\end{small}
\begin{small}
  	\vspace{-0.1cm}
\begin{equation} 
	\begin{aligned}
	&R_{t}^{dir}=c\cdot \left [ \pi -2\cdot \left | arccos\left ( \frac{\vec{v}_{i,t}\cdot \vec{goal_{i,t}}}{\left \| \vec{v}_{i,t} \right \|\cdot \left \| \vec{goal_{i,t}} \right \|} \right ) \right | \:  \right ].
	\end{aligned}
 \end{equation}
   	\vspace{-0.2cm}
\end{small}

The characters $b, c$ and following hyper-parameters $d, e, g, m, n, q$ are constant values to represent the values of the reward function, which are tunable to adjust the policy performance. We set  \begin{small}$\vec{goal}_{i,t}=d_{i,g}-\: p_{i,t}$\end{small}  goal in Eq.(9). $R_{t}^{col-s} $ and $R_{t}^{col-d} $ represent penalties for collisions with static and dynamic obstacles:
\begin{small}
  	\vspace{-0.1cm}
\begin{equation}
	R_t^{col-s} = \left \{
	\begin{aligned}
		&d, &&\text{if } \forall x\in \left [ 1,N \right ]:\left \| p_{i,t}-O_{x} \right \|\leq \,r_{i,safe}\\
		&0, &&\text{otherwise},
	\end{aligned}
\right.
\end{equation}
  	\vspace{-0.1cm}
\end{small}
\begin{scriptsize}
  	\vspace{-0.1cm}
	\begin{equation}
		R_t^{col-d} = \left \{
		\begin{aligned}
			&e,&&\text{if } \forall j\in \left [ 1,M \right ],j\neq i:\left \| \textrm{p}_{i,t}-\textrm{p}_{j,t} \right \|\leq \,r_{i,safe}+\,\textrm{r}_{j,safe}\\
			&0,&&\text{otherwise}.
		\end{aligned}
		\right.
	\end{equation}
   	\vspace{-0.2cm}
\end{scriptsize}

$R_{t}^{tim} $ is a small time penalty to facilitate reaching the target in a shorter time; $R_{t}^{goal} $ represents a reward for reaching the target, i.e.:
\begin{small}
  	\vspace{-0.2cm}
\begin{equation}
	R_t^{tim} = \left \{
	\begin{aligned}
		&g, &&\text{if time step $t$ is over } \\
		&0, &&\text{otherwise},
	\end{aligned}
	\right.
\end{equation}
  	\vspace{-0.1cm}
\end{small}
\begin{small}
  	\vspace{-0.1cm}
	\begin{equation}
		R_t^{goal} = \left \{
		\begin{aligned}
			&m, &&\text{if } \left \| p_{i,t}-\,d_{i,g} \right \|\leq q \\
			&0, &&\text{otherwise}.
		\end{aligned}
		\right.
	\end{equation}
   	\vspace{-0.2cm}
\end{small}

For norm reward $R_{t}^{norm}$, we induce agents to learn to follow social norms during training by designing a reasonably small penalty region. In particular, we will consider several common human social traffic norms (\textbf{norm-1,2,3}) in Section \uppercase\expandafter{\romannumeral4}. As in Fig.~2(a), for a differential wheeled robot-$i$, we design an asymmetric social norm penalty area $area_{i,norm} $. For a time step $t $, norm reward  $R_t^{norm} $ is as follows:

\begin{scriptsize}
  	\vspace{-0.2cm}
  	\begin{equation}
		R_t^{norm} = \left \{
		\begin{aligned}
			&n, &&\text{if } \exists j\in \left [ 1,M \right ],j\neq i:\,area_{i,norm}\bigcap area_{j,norm}\neq \varnothing  \\
			&0, &&\text{otherwise}.
		\end{aligned}
		\right.
	\end{equation}     
\end{scriptsize}

First, we discuss the situation in which the robots encounter each other head-on \textbf{(norm-1)}. With the geometric center of the agent as the midline, the width of the right region is 1.5 times that of the left region, while the front tip is in front of the right boundary of the agent, which ensures that the agents will form a preference for passing on the right when meeting head-on, as in Fig.~2(b). In the case of overtaking \textbf{(norm-2)}, the agent carries a prominent "tail" area behind the right border (as in Fig.~2(a)). When a following vehicle overtakes from the right side, the vehicle invades this area and generates a negative reward (as in Fig.~2(c)). Hence, an agent overtaking from the left side at a safe distance gains a greater reward. In the case of a crossing situation (e.g., crossroads), the two borders on the left side of the agent form a $44^{\circ} $ angle  (as in Fig.~2(a)), and the safety radius of the agent passing anti-clockwise on the right is smaller (clockwise radius $R_{1}= 0.44 \,m $, anti-clockwise radius $R_{2}= 0.31\,m $, as in Fig.~2(e),(f)). That induces the robot to form cooperative behaviors (e.g., circling in a circular loop, as in Fig.~6(a)) enabling a safer crossing. The values for our policy training in this paper are \begin{small}$b=3, c=1, d=-40, e=-15, g=-0.25, m= 80,n=-2,q=0.12m$\end{small} .

\subsection{Priority-ORCA and Algorithmic Fusion Methods}

In the previous section, we can obtain policy $\pi _{\theta }^{RL} $ that follows social rules (Task-2) and achieves cooperation by DRL with norm reward. $\pi _{\theta }^{RL} $ also makes agents better achieve pathfinding (Task-1) in non-convex static obstacle scenes without maps. However, the DRL-based navigation strategy is difficult to achieve a high success rate in complex multi-dynamic obstacle scenarios (Task-3), while the interpretability and stability of this learning strategy are insufficient. ORCA provides sufficient conditions for large-scale robotic systems to avoid collisions without communication. Knowing that ORCA guarantees obstacle avoidance for dynamic obstacles, but the algorithm assumes that all static obstacles in the scene are convex\footnote{https://gamma.cs.unc.edu/RVO2/}. For non-convex static obstacle scenes, the agent in ORCA tends to fall into local minimal regions. The ORCA algorithm cannot implement pathfinding (Task-1) properly. Meanwhile, when multiple agents meet during navigation, each agent plans its own path without cooperation, resulting in chaotic trajectories (as in Fig.~6(b)). In summary, if the advantages of DRL and ORCA are correctly combined, the fusion algorithm will overcome their respective shortcomings and significantly improve the new algorithm's performance. Next, based on policy $\pi _{\theta }^{RL} $, we will combine Priority-ORCA and propose a novel multi-agent safe reinforcement learning algorithm $\pi _{\theta }^{srl-orca} $.(shown in \textbf{Algorithm 2}).

\textbf{(1) Priority-ORCA:} In human traffic scenarios, some vehicles executing emergency tasks should have priority to pass. The integration of \textbf{norm-3} into ORCA can guarantee absolute avoidance actions, and the improved algorithm is called priority-ORCA. For two agents $i $ and $j $, the velocity obstacle $VO_{i|j}^{\tau} $ for agent $i $
at time $\tau $ can be calculated according to \cite{berg2011reciprocal}. Then we can calculate the minimum variation $\vec{u} $ (the vector from $\left ( \vec{v}^{opt}_{i}-\vec{v}^{opt}_{j} \right ) $ to the closest point on the boundary of the velocity obstacle): \begin{scriptsize}$\vec{u}= ( \underset{\vec{v}\in{\partial }VO_{i|j}^{\tau }}{\argmin} \| \vec{v}- ( \vec{v}^{opt}_{i}-\vec{v}^{opt}_{j} )  \|  )- ( \vec{v}^{opt}_{i}-\vec{v}^{opt}_{j} ) $
\end{scriptsize}, where $\vec{v}^{opt}_{i} $ is the optimization velocity and  $\vec{n} $ denotes the outward normal of the boundary of  $VO_{i|j}^{\tau } $ at point  $\left ( \vec{v}^{opt}_{i}-\vec{v}^{opt}_{j} \right ) +\vec{u} $. Unlike ORCA, which shares the avoidance volume equally, Priority-ORCA considers that the avoidance variation $\vec{u} $ is allocated among agents in proportion to their priorities $pr_{i} $ and $pr_{j} $ (\textbf{norm-3}). So vehicles with higher priorities take less responsibility for avoidance to obtain higher speeds and smoother trajectories for urgent tasks. As in Fig.~3(a), the set of collision-free velocities $ORCA_{i|j}^{\tau-prior } $ for agent $i $ relative to robot $j $ can geometrically be constructed from  $VO_{i|j}^{\tau} $,

\begin{small}
  	\vspace{-0.3cm}
	\begin{equation}
	ORCA_{i|j}^{\tau-prior }=\left \{ \vec{v}|\vec{v}-\left ( \vec{v}_{i}^{opt} +\frac{pr_{j}}{pr_{i}+\,pr_{j}} \cdot \vec{u}\right )\cdot \vec{n} \geq  0\right \},
	\end{equation}
\end{small}

\noindent where $ORCA_{i|j}^{\tau-prior } $ (red region in Fig.~3(a)) has an avoidance offset in the direction of  $\vec{n} $ with respect to $ORCA_{i|j}^{\tau } $ (purple region in Fig.~3(a)). The set of collision-free velocities $ORCA^{\tau-prior }_{i} $ relative to all agents can be given as

\begin{small}
  	\vspace{-0.2cm}
	\begin{equation}
		ORCA^{\tau-prior}_{i}=S_{AHV_{i}}^{RL}\cap \bigcap_{j\neq i}ORCA_{i|j}^{\tau-prior }.
	\end{equation}
   	\vspace{-0.4cm}
\end{small}

\begin{figure}[!t]
	\centering
	\vspace{-0.0cm}	\includegraphics[width=3.4in]{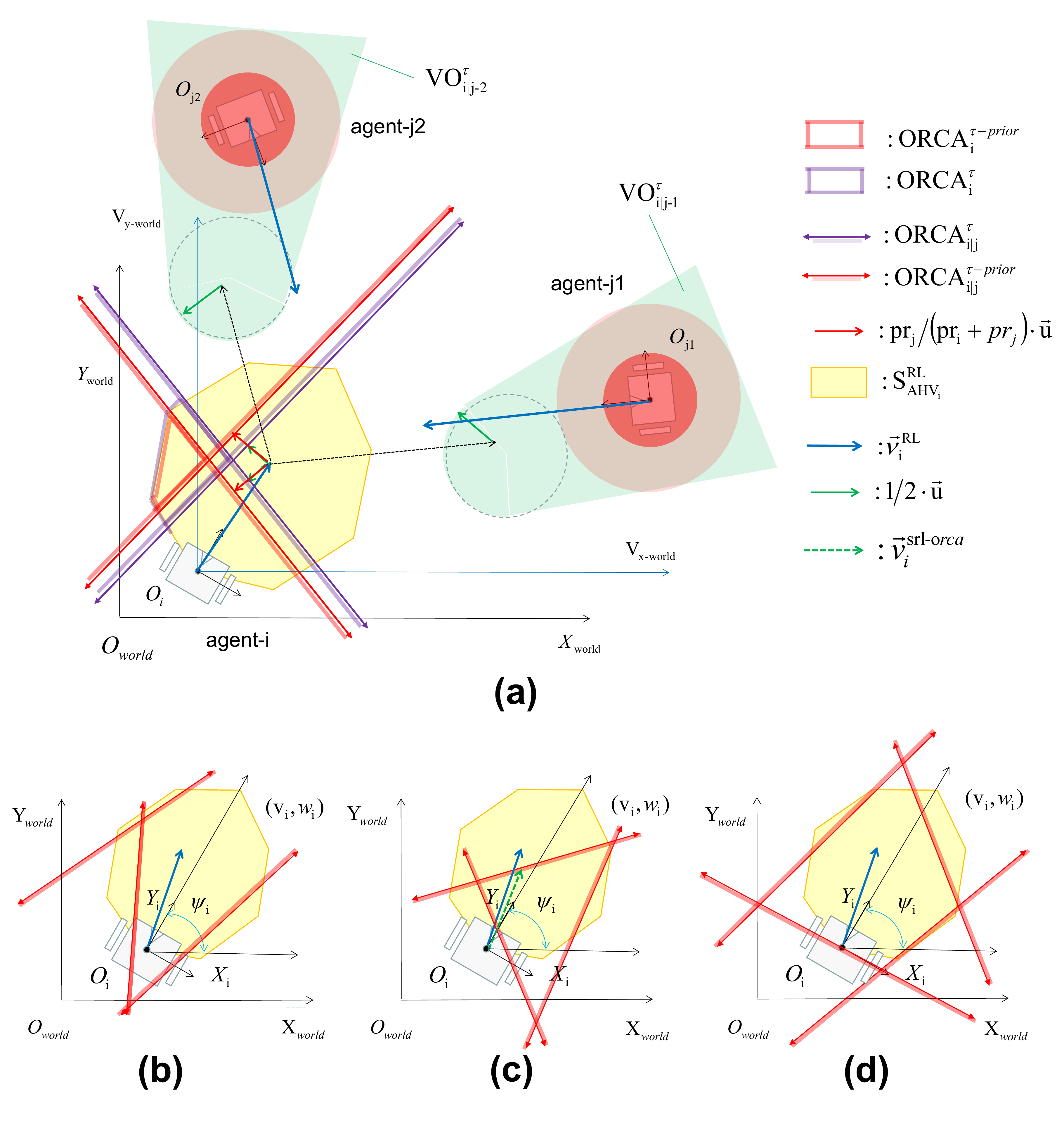}
 	\vspace{-0.7cm}
	\caption{(a) Calculate the set $ORCA^{\tau-prior}_{i}$ based on $VO_{i|j}^{\tau} $, $pr_{i} $, and $pr_{j} $; (b) Case-1; (c) Case-2; (d) Case-3. Case-1 is a safe scenario, and we directly use the policy $\pi _{\theta }^{RL} $ of DRL to achieve navigation. When unsafe situations are encountered (Case-2 and Case-3), we use the advice provided by external knowledge (ORCA) to modify the policy $\pi _{\theta }^{RL} $ of DRL to avoid catastrophic collisions.}
  	\vspace{-0.5cm}
	\label{fig7}
\end{figure}

\textbf{(2) The design of SRL-ORCA:} 
We already obtains the navigation policy $\pi _{\theta }^{RL} $ adapted to the complex non-convex scenarios (with social \textbf{norm-1,2}) and the priority permitted speed set $ORCA^{\tau-prior }_{i} $ (with \textbf{norm-3}), respectively. Policy $\pi _{\theta }^{RL} $ can achieve better results in purely static scenarios, but due to the high dimensionality and complexity of the dynamic obstacle avoidance problem, it is difficult to achieve a high success rate in scenarios mixed with a large number of dynamic obstacles and static obstacles. Then we will incorporate DRL-part and OCRA-part in the same framework to obtain a better fusion algorithm SRL-ORCA. In this work, two sets of algorithms operate in parallel (as in Fig.~1). We design a switching method to implement multi-agent safe reinforcement learning, as in \textbf{Algorithm 2}. The fusion algorithm must better implement the three tasks proposed in Section-II and follow the three social rules in Section-III.

\begin{figure}[!htb]
  	\vspace{-0.4cm}
	\label{alg:Execution}
	\removelatexerror
	\begin{algorithm}[H]
		\caption{The execution algorithm of SRL-ORCA}
		\begin{algorithmic}[1]
			\Require Policy $\pi _{\theta }^{RL} $. Agents $i\in \left [ 1,M \right ] $ provide with: set $S_{AHV_{i}}^{RL} $, state $s_{i,t}$.
			\Ensure  Actions $a_{i,t}^{srl-orca}=\vec{v}_{i,t}^{srl-orca}=\left [ v_{i,t},\omega_{i,t} \right ] $.
			\Loop
			\For{$i\in \left \{ 1,...,M \right \} $}
			\State Obtain action $\vec{v}^{RL}_{i} $ based on policy $\pi _{\theta }^{RL}  $.
			\State Construct $ORCA^{\tau-prior}_{i}$ following Eq.(15)(16).
			
            \If{\begin{small}$ORCA^{\tau-prior}_{i} \neq \emptyset,\, \vec{v}^{RL}_{i}\in \,ORCA^{\tau-prior}_{i} $\end{small}}
			\State \textbf{Case-1:} $\vec{v}^{srl-orca}_{i}=\,\vec{v}^{RL}_{i} $.			
			\ElsIf{\begin{scriptsize}
			    $ORCA^{\tau-prior}_{i} \neq \emptyset, \vec{v}^{RL}_{i}\notin ORCA^{\tau-prior}_{i} $
			\end{scriptsize}}
			\State \textbf{Case-2:} Compute $\vec{v}^{srl-orca}_{i} $ by Eq.(17).						\ElsIf{\begin{small}$ORCA^{\tau-prior}_{i} = \emptyset,\, \vec{v}^{RL}_{i}\in S_{AHV_{i}}^{RL} $\end{small}}
			\State \textbf{Case-3:} Compute $\vec{v}^{srl-orca}_{i} $ by Eq.(18).
			\EndIf
			
			\State \textbf{Output:} Actions  $\vec{v}^{srl-orca}_{i}$ for agent $i $.
			\EndFor
			\EndLoop
			
		\end{algorithmic}
	\end{algorithm}
   	\vspace{-0.4cm}
\end{figure}

In the absence of a global map, the agent's learned policy $\pi _{\theta }^{RL} $ can react based on current state $s_{i,t} $ and achieve navigation. We select the optimal speed for agent $i $ as $\vec{v}_{i}^{opt}= \,\vec{v}^{RL}_{i}$ by policy $\pi _{\theta }^{RL} $. Here, the optimal speed  $\vec{v}_{i}^{opt} $ is a time-varying speed that adapts automatically to the scene. When dealing with non-convex static areas, policy $\pi _{\theta }^{RL} $ can bypass non-convex obstacle areas rather than easily falling into local minimal regions (as in Fig.~7(a),(b)). DRL implements pathfinding (Task-1), overcoming the problem that ORCA cannot cope with non-convex obstacles. When agents encounter dynamic obstacles, DRL provides interactions that obey the norms (\textbf{norm-1, 2}) and Priority-ORCA provides \textbf{norm-3} (Task-2), which avoids uncooperative congestion and chaotic trajectories in ORCA (as in Fig.~6(a),(b)).

In scenarios with high robot density, the behavior of collision avoidance requires very precise and robust actions of the controller. Yet the neural network trained by DRL (policy $\pi _{\theta }^{RL} $) has poor generalization ability to perfectly cope with dynamic obstacle avoidance. We use ORCA as an external knowledge to implement safe reinforcement learning in the specific problem of mapless navigation \cite{garcia2015comprehensive}. When unsafe situations are encountered (Case-2 and Case-3), external knowledge provides safety advice (set $ORCA^{\tau-prior }_{i} $) to the DRL, which can optimize policy to avoid unsafe actions (e.g. catastrophic collisions) \cite{maclin1996creating}. It can absolutely guarantee the safety of dynamic or static collision avoidance (Task-3), and improve the shortcomings of dynamic obstacle avoidance for policy $\pi _{\theta }^{RL} $. Based on speed $\vec{v}^{RL}_{i} $ and safety advice $ORCA^{\tau-prior }_{i} $, we design the switching method of SRL-ORCA and discuss three scenarios:

\textbf{Case-1:} $ORCA^{\tau-prior }_{i} \neq \emptyset,\, \vec{v}^{RL}_{i}\in \,ORCA^{\tau-prior }_{i} $

This case indicates that speed $\vec{v}^{RL}_{i} $ is the safe speed that fulfills the priority rule and speed $\vec{v}^{RL}_{i} $ will guarantee no collision with any other agents in time $\tau $. The action of policy $\vec{v}^{RL}_{i} $ is within the scope of safety advice from external knowledge (ORCA). The action $\vec{v}^{RL}_{i} $ is safe, and we can let $\vec{v}^{srl-orca}_{i}=\,\vec{v}^{RL}_{i} $, as in Fig.~3(b). Velocity $\vec{v}^{srl-orca}_{i} $ is the speed of the next step that agent $i $ eventually executes.

\textbf{Case-2:} $ORCA^{\tau-prior }_{i} \neq \emptyset,\, \vec{v}^{RL}_{i}\notin \,ORCA^{\tau-prior }_{i} $

Velocity $\vec{v}^{RL}_{i} $ cannot guarantee the priority rule or no collision. We optimize the actions of policy $\vec{v}^{RL}_{i} $ based on the advice provided by external knowledge (ORCA). Since the set \begin{small}$ORCA^{\tau-prior }_{i} $\end{small} is a convex region, we can use a 2D linear optimization function to choose a new velocity $\vec{v}^{srl-orca}_{i} $ in the set \begin{small}$ORCA^{\tau-prior }_{i} $\end{small} that is closest to the velocity $\vec{v}^{RL}_{i} $. As in Fig.~3(c), velocity $\vec{v}^{srl-orca}_{i} $ is given by:
\begin{small}
  	\vspace{-0.2cm}
	\begin{equation}
	\begin{split}
		&\vec{v}^{srl-orca}_{i}=\argmin\left \| \vec{v}-\, \vec{v}^{RL}_{i}\right \|\\
		&s.t.\:\vec{v}\in\, ORCA^{\tau-prior }_{i}.
	\end{split}
	\end{equation}
   	\vspace{-0.2cm}
\end{small}

\textbf{Case-3:} $ORCA^{\tau-prior }_{i} = \emptyset,\, \vec{v}^{RL}_{i}\in S_{AHV_{i}}^{RL} $
	
This is the case with a high density of dynamic obstacles, therefore the set \begin{small}$ORCA^{\tau-prior }_{i} $\end{small} is empty. Here, the linear programming in Equation (17) is not feasible. 
This is a dangerous situation prone to collision and the controller requires very precise actions to guarantee safety. SRL-ORCA temporarily interrupts action $\vec{v}^{RL}_{i} $ of the DRL and executes the action given by the external knowledge (ORCA). To ensure that there are no collisions between agents and they continue to move without local deadlock, we still use the 3D linear programming of ORCA\cite{berg2011reciprocal} within the set $S_{AHV_{i}}^{RL} $, as in Equation (18). Let $d_{i|j}\left ( \vec{v} \right ) $ denote the signed distance of velocity $\vec{v} $ to the edge of the half-plane \begin{small}$ORCA_{i|j}^{\tau-prior } $\end{small}. As in Fig.~3(d), we select the velocity $\vec{v}^{srl-orca}_{i} $ that minimizes the maximum $d_{i|j}\left ( \vec{v} \right ) $ induced by the other agents:
\begin{small}
  	\vspace{-0.1cm}
	\begin{equation}
		\begin{split}
		&\vec{v}^{srl-orca}_{i}=\argmin\:  \underset{i\neq j}{\max}\:  d_{i|j}\left ( \vec{v} \right )\\
		&s.t.\:\vec{v}\in S_{AHV_{i}}^{RL}.
		\end{split}
	\end{equation}
   	\vspace{-0.3cm}
\end{small}

During the learning process in DRL with a safe social distance penalty (Eq.(14)), policy $\pi _{\theta }^{RL} $ will keep the agents at a safe distance from each other so that high-density situation (Case-3) is less likely to occur. Thereby they are less prone to slow down due to crowding (as in Fig.~6(a),(b)). Once the robots have crossed the crowded area (Case-3), we resume the execution of policy $\pi _{\theta }^{RL} $ of DRL, which will continue to navigate the robots toward their targets.

In summary, SRL-ORCA effectively implements safe reinforcement learning \cite{garcia2015comprehensive}\cite{maclin1996creating}. All of Task-1,2,3 can be better implemented, and the fusion algorithm will achieve stronger performance than DRL or ORCA. We will validate the navigation performance of SRL-ORCA in Section-\uppercase\expandafter{\romannumeral5}.

\begin{figure}[htbp]
	\centering
	\vspace{-0.0cm}
	\subfigure[]{
		\begin{minipage}[b]{0.49\textwidth}			\includegraphics[width=1\textwidth]{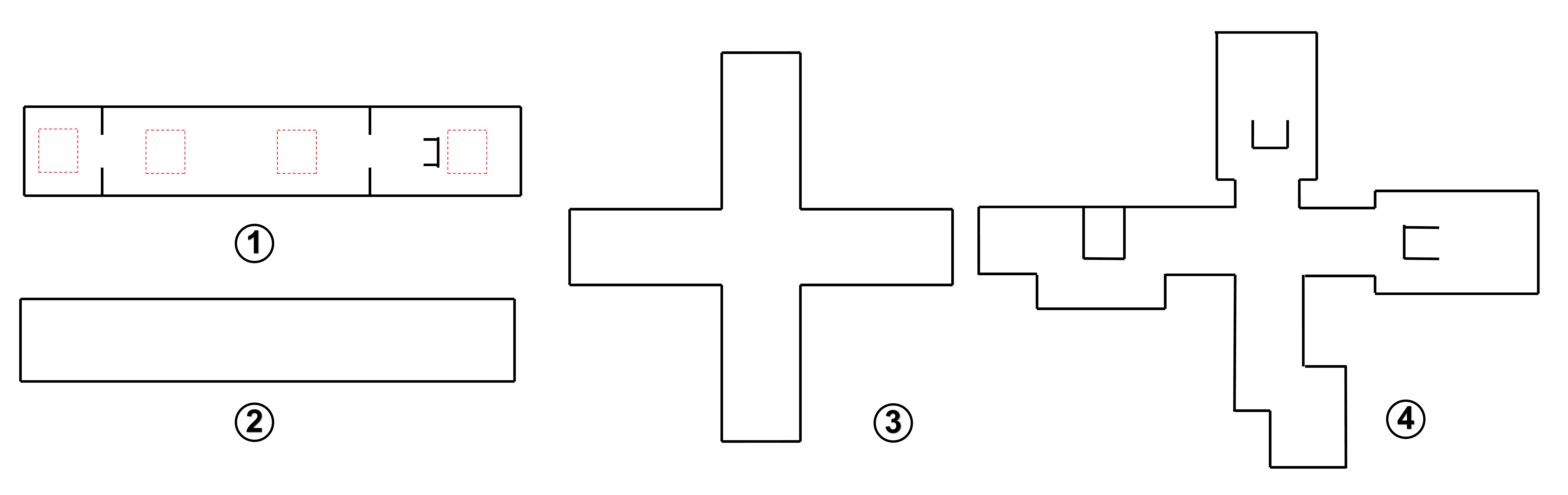}
		\end{minipage}
	}
	\vspace{-0.3cm}	
	\subfigure[]{
		\begin{minipage}[b]{0.49\textwidth}			\includegraphics[width=1\textwidth]{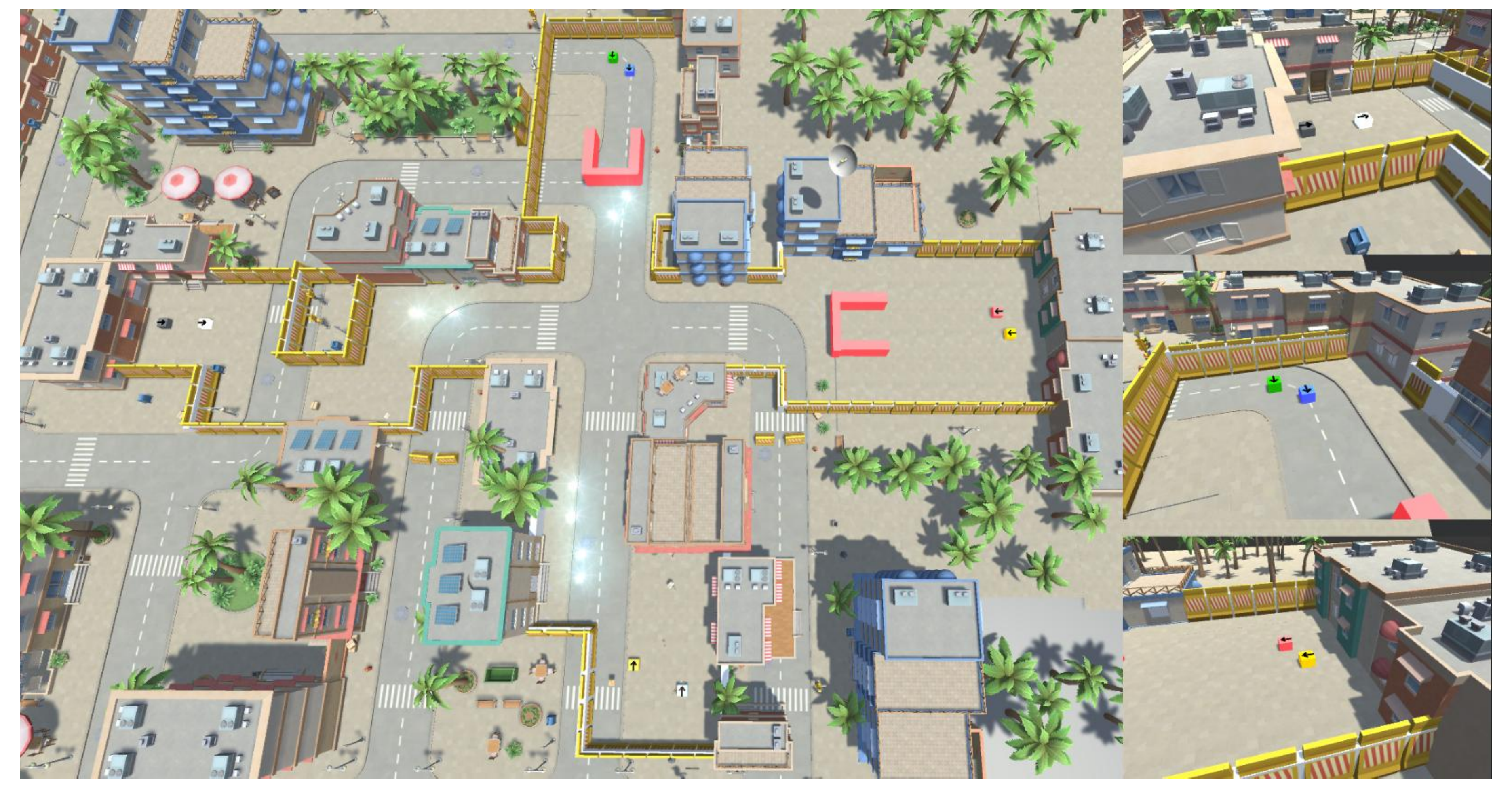}
		\end{minipage}
	}	
 	\vspace{-0.3cm}	
	\subfigure[]{
		\begin{minipage}[b]{0.49\textwidth}			\includegraphics[width=1\textwidth]{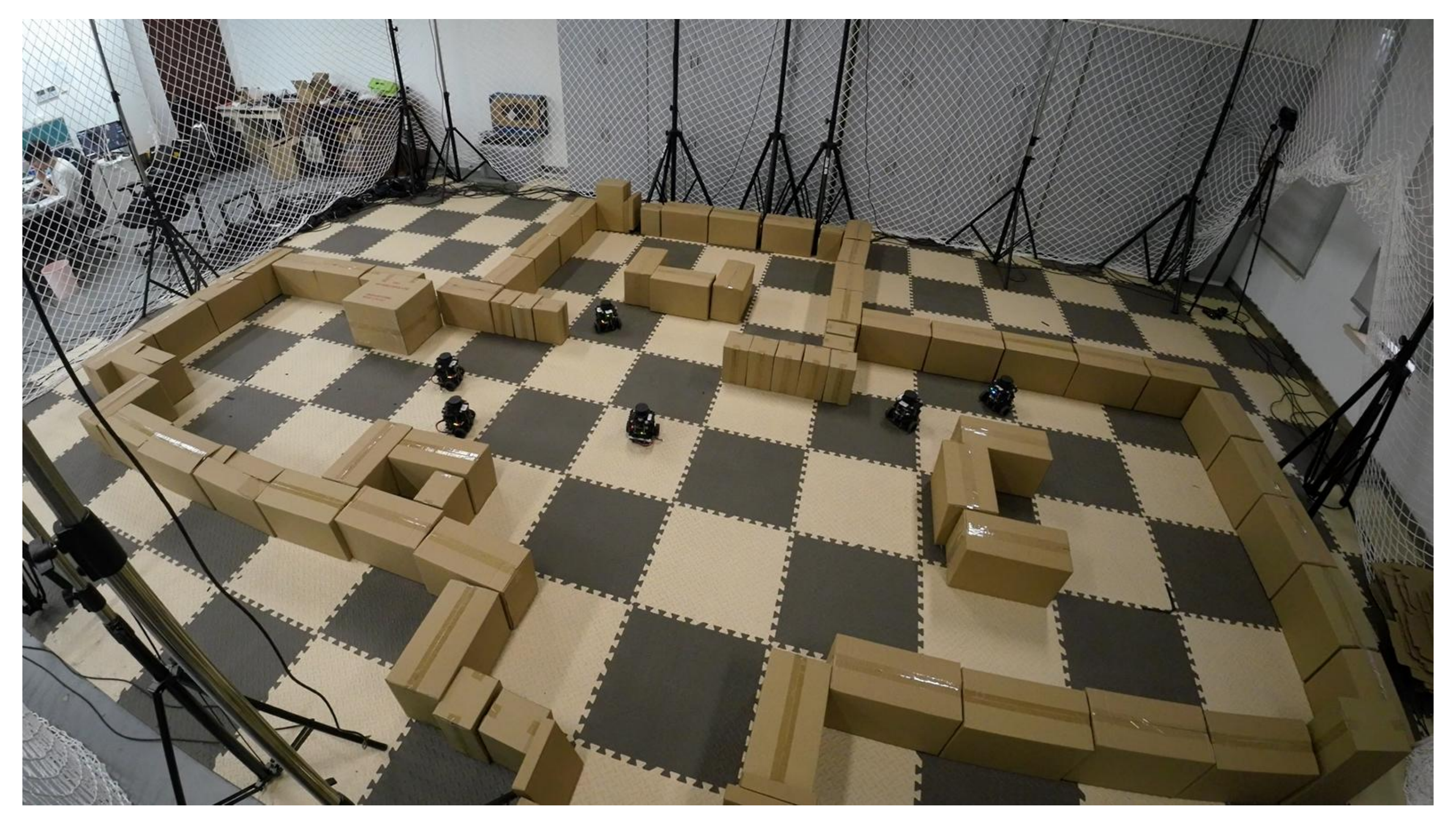}
		\end{minipage}
	}	
	\caption{(a) the structure of testing scenes for Scenarios 1-4; (b) shows a simulated town with multiple intersections and many obstacles in Unity, and agents are trained in it. The left side is the simulation scene. On the right side, we provide three side camera views to observe the agent's motion. (c) is a real robot swarm (6 Turtlebot3 robots) for experiments.}
	\label{fig9}
   	\vspace{-0.4cm}
\end{figure}

\section{RESULTS}

This section will quantitatively evaluate the navigation performance of four algorithms, SRL-ORCA, NH-ORCA, DRL, and Hybrid-RL\cite{fan2020distributed}. SRL-ORCA is implemented in Pytorch, trained in Gazebo, visualized in Unity, and transferred to a real robot swarm (Turtlebot3 robots) for experiments. Fig.~4(a) shows the structure of the four scenarios for testing, and Fig.~4(b) shows a simulated town with multiple intersections and many obstacles, and agents are trained in it. The training process is executed on a desktop with CPU i9-11900K and GPU Nvidia GTX 3070Ti. 
The training of the algorithm takes 5 million steps. We set \begin{small}$ v_{t}\in \left ( 0.01,0.20 \right ) m/s,\:  w_{t}\in \left ( -2.5,2.5 \right ) rad/s $\end{small}, time step \begin{small}$t=0.2s $\end{small}, curiosity strength \begin{small}$\delta =0.01 $\end{small}. Three metrics are introduced to evaluate the performance of the navigation algorithm as follows: 

\begin{enumerate}
	\item{Success rate ($\bar{\chi } $): }The percentage of episodes in which robots successfully reach their target without collisions.
	\item{Average cycle time ($\bar{t}$): }The average cycle time spent by all arrived robots in successful episodes. Fewer cycle time indicates more efficient navigation.
	\item{Social rule success rate ($\bar{\chi_{s} }$): }The proportion of successful events in which the robot follows social rules. 
\end{enumerate}


\subsection{Performance of Reinforcement Learning Policy $\pi _{\theta}^{RL} $}

We compare the navigation performance of the curiosity-driven policy $\pi _{\theta }^{RL-cd} $ and the curiosity-free policy $\pi _{\theta }^{RL-cf} $ under Scenario-4 in Section \uppercase\expandafter{\romannumeral5}-B. Here, Scenario-4 is a complex non-convex static scene containing dynamic obstacles. The success rates of $\pi _{\theta }^{RL-cd} $ and $\pi _{\theta }^{RL-cf} $ are compared in both the sparse and dense reward condition (as in Fig.~5).

\begin{figure}[htbp]
	\centering
	\vspace{-0.3cm}
\centering
\subfigure[]{
\begin{minipage}[b]{1\linewidth}
\centering
\includegraphics[width=1\textwidth]{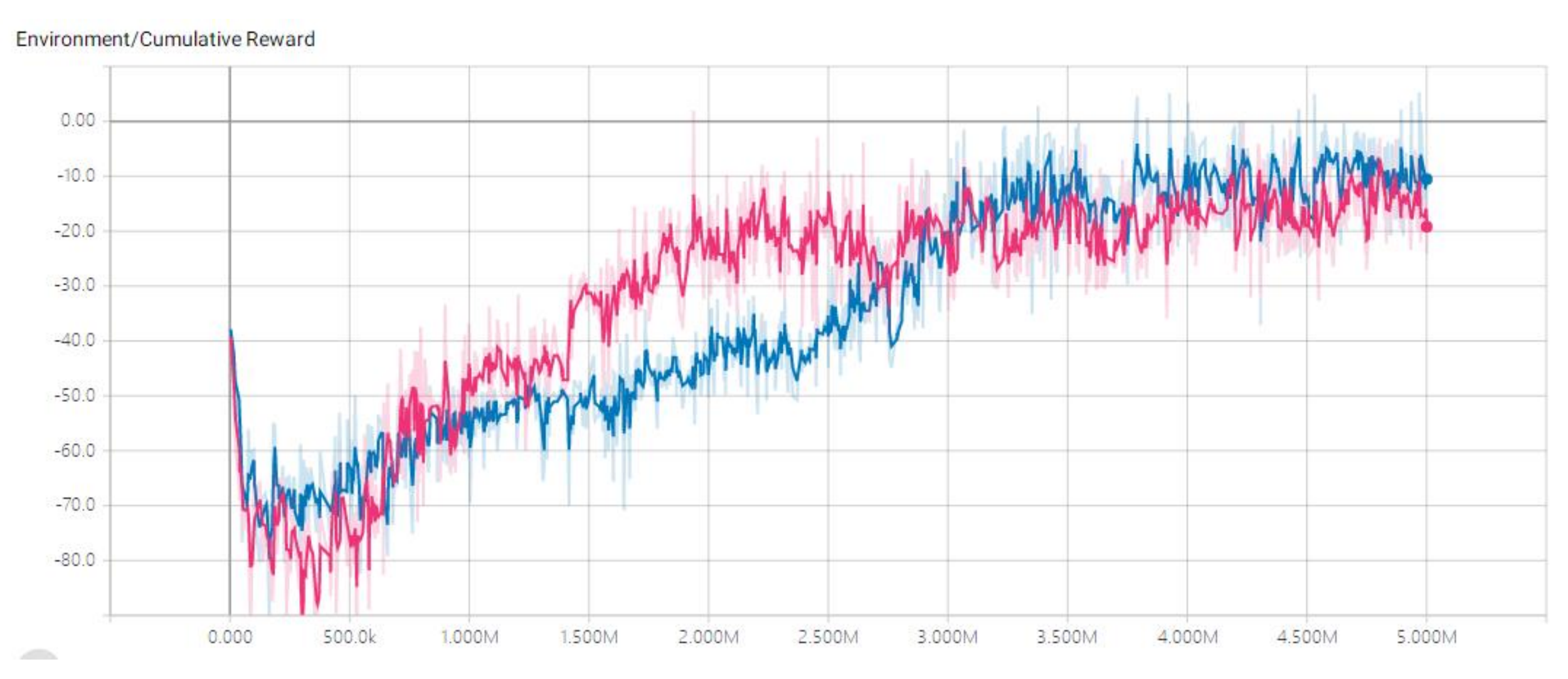}
\end{minipage}%
}%
 	\vspace{-0.3cm}
\subfigure[]{
\begin{minipage}[b]{1\linewidth}
\centering
\includegraphics[width=1\textwidth]{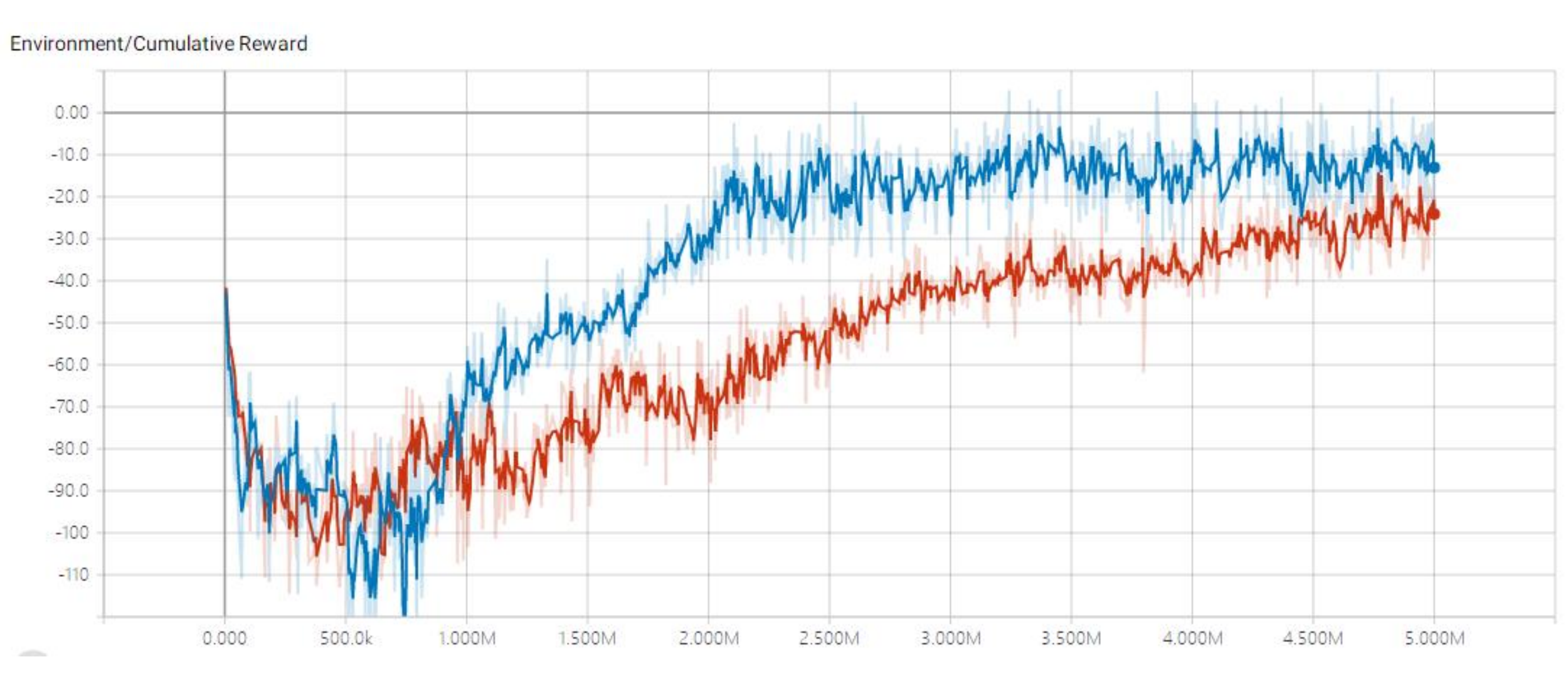}
\end{minipage}%
}%

\centering
\vspace{-0.1cm}
\caption{Environmental reward curve of SRL-ORCA algorithm in Scenario-4. (a) is under sparse rewards (moving forward reward $R _{t }^{mf}=0$ and direction reward $R _{t }^{dir}=0$), (b) is under normal rewards (moving forward reward $R _{t }^{mf}$ and direction reward $R _{t }^{dir}$ set as in Section \uppercase\expandafter{\romannumeral4}-A(2)).}
	\label{fig10}
   	\vspace{-0.2cm}
\end{figure}

\textbf{(1) Sparse rewards:} From Fig.~5(a) we can see that the reward of $\pi _{\theta }^{RL-cd} $ (blue line) is smaller than that of $\pi _{\theta }^{RL-cf} $ (pink line) when training steps is less than 2.7 million under the sparse reward condition. At this time, the agents of policy $\pi _{\theta }^{RL-cd} $ are more actively exploring the environment, thus the reward of $\pi _{\theta }^{RL-cd} $ is smaller than that of $\pi _{\theta }^{RL-cf} $. After the steps exceed 2.7 million, policy $\pi _{\theta }^{RL-cd} $ obtains greater external rewards due to deeper exploration of the environment and more efficient utilization of environmental information. Ultimately, $\pi _{\theta }^{RL-cd} $ obviously achieved a better success rate by adding the curiosity module (as in \textbf{Table I}, $\bar{\chi }^{drl-cd}=75.0\%$ and $\bar{\chi }^{drl-cf}=37.2\%$).

\textbf{(2) Normal rewards:} Since external rewards are more intensive (as in Section \uppercase\expandafter{\romannumeral4}-A(2)), the agents explore the environment faster than that of the sparse reward situation under the combined effect of external and curiosity rewards. Consequently, policy $\pi _{\theta }^{RL-cd} $ (blue line) outperforms policy $\pi _{\theta }^{RL-cf} $ (orange line) at 0.9 million training steps in Fig.~5(b).
As the steps increase from 0.9 to 5 million, policy $\pi _{\theta }^{RL-cd} $ is more fully optimized and eventually navigates better with greater external rewards (as in \textbf{Table \uppercase\expandafter{\romannumeral1}}, $\bar{\chi }^{drl-cd}=84.0\%$ and $\bar{\chi }^{drl-cf}=75.2\%$).

The above results demonstrate that curiosity-driven reinforcement learning has acquired more effective navigation policies and improved the speed of training.

\begin{table}
  	\vspace{-0.1cm}
\centering
\caption{Experimental results from Scenario-1 to Scenario-4}
\resizebox{\linewidth}{!}{

\begin{tblr}{
  cells = {c},
  cell{2}{1} = {r=3}{},
  cell{2}{2} = {r=3}{},
  cell{5}{1} = {r=3}{},
  cell{5}{2} = {r=3}{},
  cell{8}{1} = {r=5}{},
  cell{8}{2} = {r=5}{},
  cell{13}{1} = {r=5}{},
  cell{13}{2} = {r=5}{},
  hline{1-2,5,8,13,18} = {-}{},
}
{Number\\of robots} & Scenario                 & Method    & {Success \\Rate(\%)} & {SR-\\success \\rate(\%)} & {Col-With\\~robots(\%)} & {Col-with\\~obstacles(\%)} & {Time \\Out(\%)} & {Rotate in \\place(\%)} & {Average \\Time(s)} \\
2                   & 1-Passing                & DRL-cd    & 96.4~                & 92.8~                     & 2.4~                    & 1.2~                       & 0.0~             & 0.0~                    & 19.6                \\
                    &                          & NH-ORCA   & 55.7~                & 48.4~                     & 0.0~                    & 0.0~                       & 44.3~            & 0.0~                    & fail                \\
                    &                          & SRL-ORCA  & \textbf{97.3~}       & \textbf{94.6~}            & 0.0~                    & 2.7~                       & 0.0~             & 0.0~                    & 20.6                \\
2                   & 2-Overtaking             & DRL-cd    & \textbf{100.0~}      & \textbf{100.0~}           & 0.0~                    & 0.0~                       & 0.0~             & 0.0~                    & 25.4                \\
                    &                          & NH-ORCA   & \textbf{100.0~}      & 49.8~                     & 0.0~                    & 0.0~                       & 0.0~             & 0.0~                    & 24.4                \\
                    &                          & SRL-ORCA  & \textbf{100.0~}      & \textbf{100.0~}           & 0.0~                    & 0.0~                       & 0.0~             & 0.0~                    & \textbf{23.4}       \\
8                   & 3-Crossroad              & DRL-cf    & 54.7~                & -                         & 38.0~                   & 5.3~                       & 1.3~             & 0.7~                    & fail                \\
                    &                          & DRL-cd    & 58.0~                & -                         & 38.7~                   & 1.3~                       & 2.0~             & 0.0~                    & fail                \\
                    &                          & Hybrid-RL & 73.7~                & -                         & 23.9~                   & 2.4~                       & 0.0~             & 0.0~                    & fail                \\
                    &                          & NH-ORCA   & 100.0~               & -                         & 0.0~                    & 0.0~                       & 0.0~             & 0.0~                    & 31.90~              \\
                    &                          & SRL-ORCA  & 90.7~                & -                         & 2.7~                    & 3.3~                       & 1.3~             & 2.0~                    & \textbf{27.95~}     \\
8                   & {4-N-C-O \\Intersection} & DRL-cf    & 75.2~                & -                         & 21.5~                   & 3.3~                       & 0.0~             & 0.0~                    & fail                \\
                    &                          & DRL-cd    & 84.0~                & -                         & 10.7~                   & 5.3~                       & 0.0~             & 0.0~                    & fail                \\
                    &                          & Hybrid-RL & 82.7~                & -                         & 14.6~                   & 2.2~                       & 0.5~             & 0.0~                    & fail                \\
                    &                          & NH-ORCA   & 25.0~                & -                         & 0.0~                    & 0.0~                       & 75.0~            & 0.0~                    & fail                \\
                    &                          & SRL-ORCA  & \textbf{90.0~}       & -                         & 0.0~                    & 4.7~                       & 2.0~             & 3.3~                    & \textbf{33.65}      
\end{tblr}

}
  	\vspace{-0.3cm}
\end{table}

\subsection{Experiments from Scenario-1 to Scenario-4}

To sufficiently compare the performance of these algorithms, both SRL-ORCA and DRL use the same neural network trained with the same reward function as the controller. We refer the readers to our supplementary experimental videos for more intuitive comparisons.

\textbf{Scenario-1 and Scenario-2:} For Scenario-1, two robots are randomly generated from the central red areas and travel head-on in the corridor. They need to avoid some non-convex/convex static obstacles and then reach the targets that randomly appear in the red areas at both edges. SRL-ORCA follows the rule of passing on the right side (\textbf{norm-1}) and maintains an adequate safety distance, while NH-ORCA passes on the right or left side randomly. Meanwhile, since NH-ORCA cannot deal with non-convex obstacles, the green robot is stuck in a simple local minimal region. From \textbf{Table \uppercase\expandafter{\romannumeral1}}, SRL-ORCA achieves a higher SR(social rule) success rate ($\bar{\chi }_{s}^{srl-orca}= 94.6\%$) and is more disciplined compared to the NH-ORCA or DRL ($\bar{\chi }_{s}^{nh-orca}= 48.4\%$, $\bar{\chi }_{s}^{drl-cd}= 92.8\%$). When the two robots meet, the blue robot can urgently avoid the green robot, because the green one has priority (\textbf{norm-3}). As a result, the green robot obtains a higher speed and a smoother trajectory. For Scenario-2, the red robot overtakes the yellow robot, and then both of them reach their destinations in the corridor. SRL-ORCA shows that the red robot overtakes the previous yellow robot from the left side (\textbf{norm-2}) and maintains a proper overtaking distance, while NH-ORCA still randomly overtakes from the left or right side. Similarly, \textbf{Table \uppercase\expandafter{\romannumeral1}} indicates that SRL-ORCA has a higher social rule success rate than NH-ORCA ($\bar{\chi }_{s}^{srl-orca}= 100.0\%, \bar{\chi }_{s}^{nh-orca}=48.4\%$) in the overtaking scenario. In summary, SRL-ORCA implements \textbf{norm-1,2,3}. Results for Scenario-1,2 demonstrate that SRL-ORCA achieves the task of obeying specific social norms (Task-2).

\begin{figure*}[htbp]
	\centering
	\vspace{-0.5cm}
\centering
\subfigure[]{
\begin{minipage}[b]{0.33\linewidth}
\centering
\includegraphics[width=0.9\textwidth]{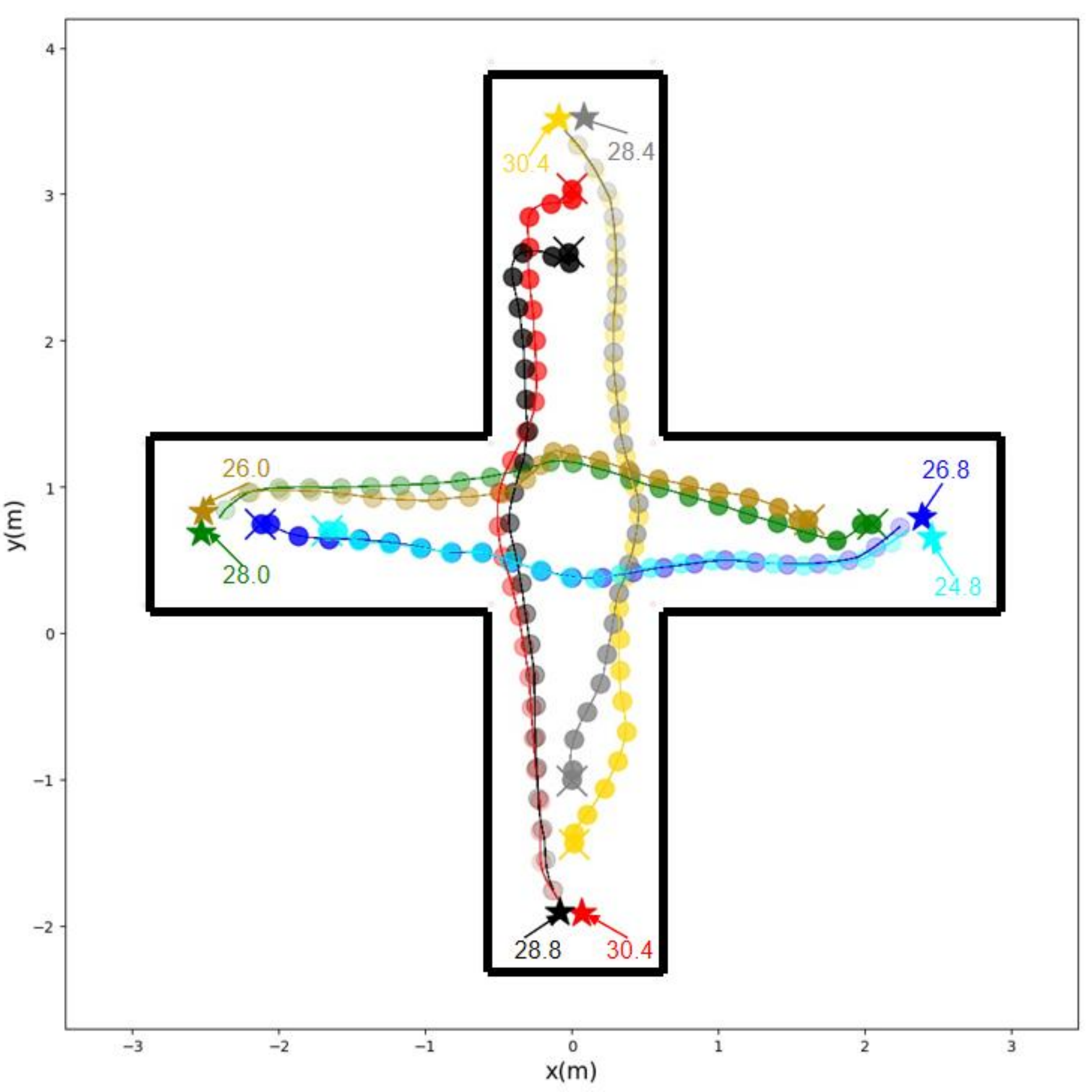}
\end{minipage}%
}%
\subfigure[]{
\begin{minipage}[b]{0.33\linewidth}
\centering
\includegraphics[width=0.9\textwidth]{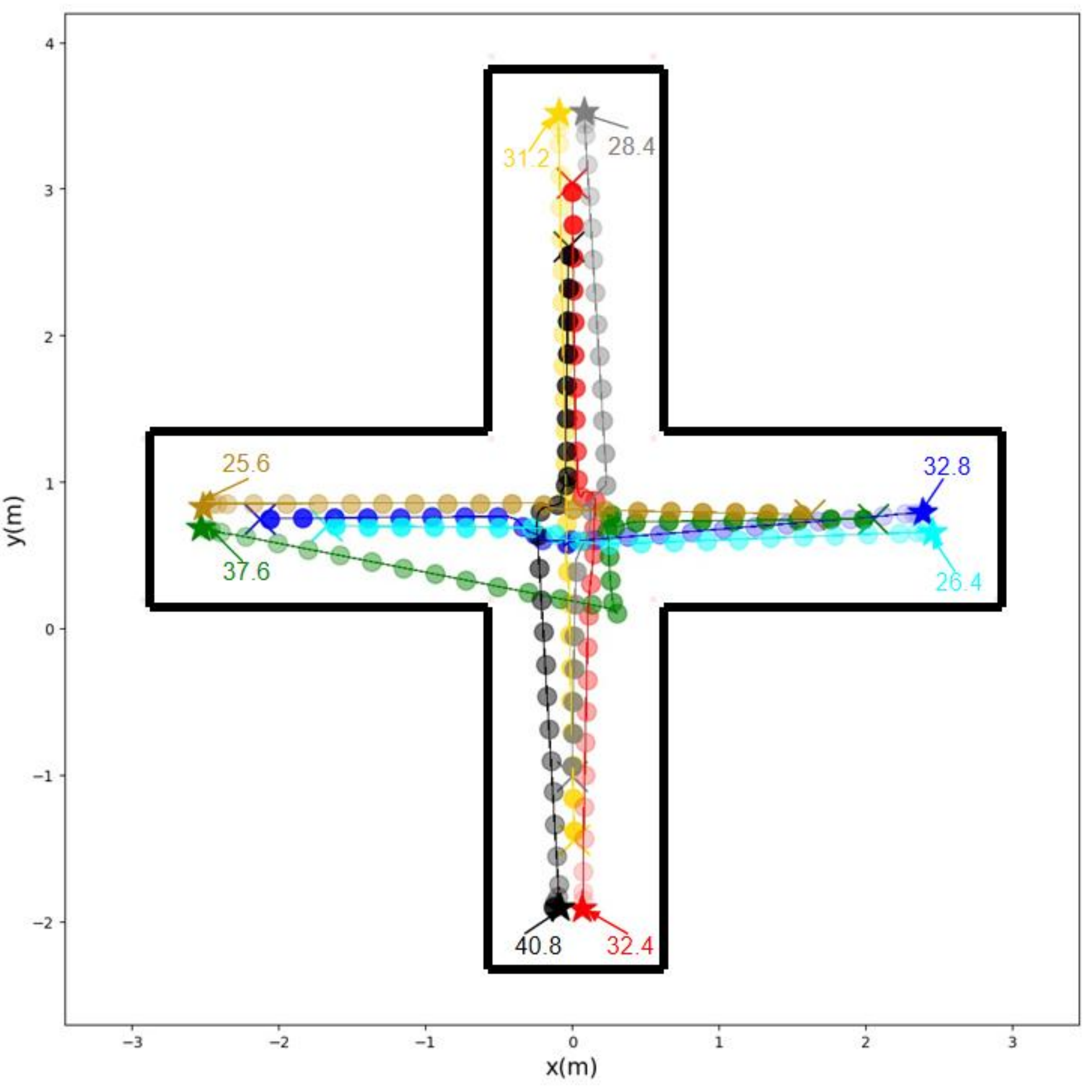}
\end{minipage}%
}%
\subfigure[]{
\begin{minipage}[b]{0.33\linewidth}
\centering
\includegraphics[width=0.9\textwidth]{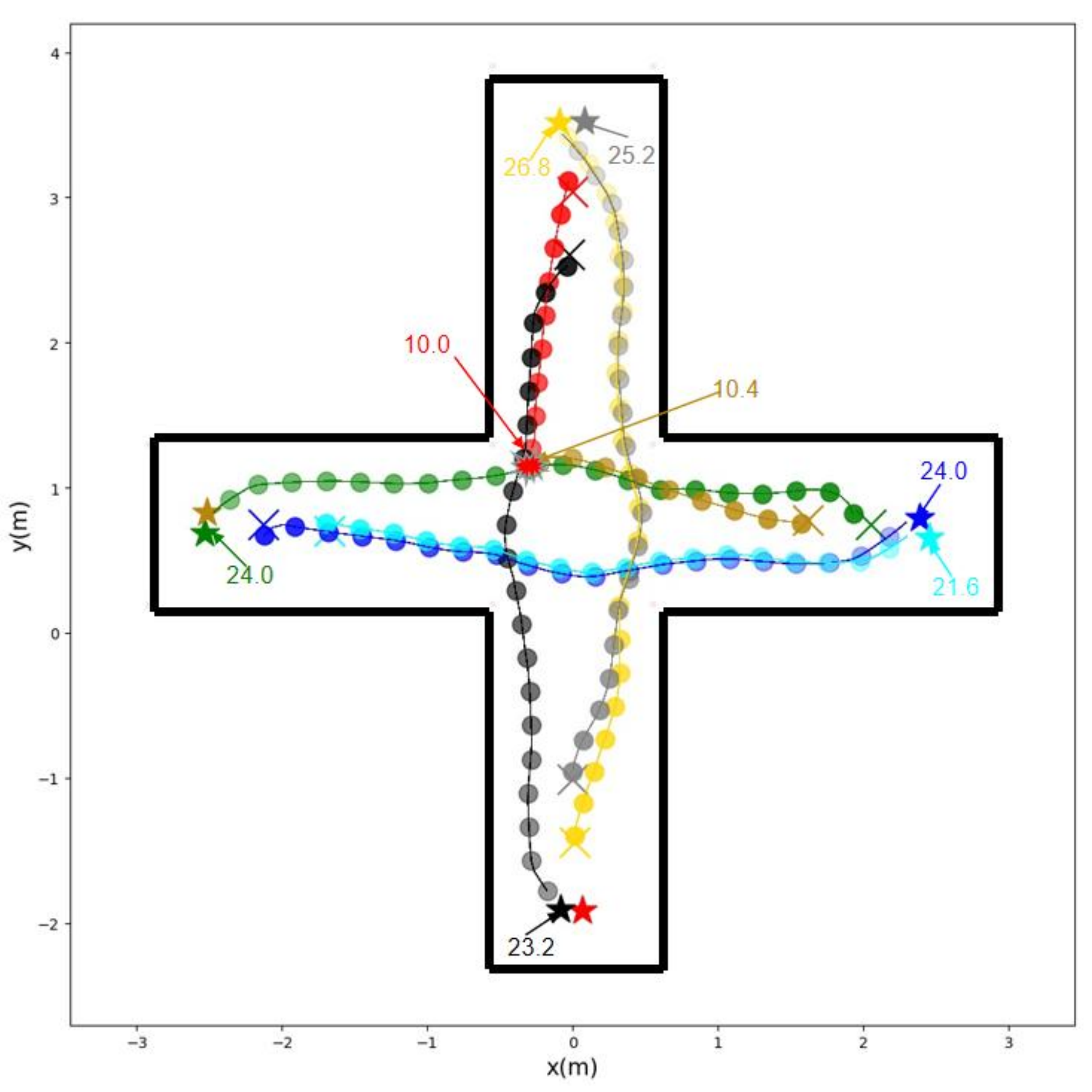}
\end{minipage}
}%

\centering

\caption{Scenario-3: 8 robots each start from 4 different directions and cross the intersection to reach their destinations. (a) shows that SRL-ORCA can coordinate robots well to pass on the right and avoid oncoming vehicles. (b) shows that NH-ORCA produces a chaotic interlaced trajectory in the intersection area. (c) shows that only 6 of the robots using DRL reach their targets safely, while the other 2 robots collide with each other.}
	\label{fig13}
   	\vspace{-0.3cm}
\end{figure*}

\textbf{Scenario-3: }As in Fig.~6, 8 robots cross the intersection. In Fig.~6(a), it can be illustrated that SRL-ORCA can coordinate robots well to pass on the right and avoid oncoming vehicles. At the intersection multiple robots spontaneously form a loop, thus avoiding collisions without reducing the speed too much. 
In contrast, NH-ORCA is not applicable to social cooperation, which produces a chaotic interlaced trajectory at the intersection 
(as in Fig.~6(b)).
The speed of each robot is very low in order to achieve dynamic obstacle avoidance. Thus, \textbf{Table \uppercase\expandafter{\romannumeral1}} shows that SRL-ORCA arrives $14.1\% $ earlier than NH-ORCA and has a better global trajectory. Compared to NH-ORCA, SRL-ORCA effectively increases the average speed by enabling cooperation and social norms. This results in a faster and smoother flow of moving robots in scenarios with high robot density.

From Fig.~6(c), although there is cooperation among the robots using DRL, only 6 robots reach their destinations safely, while 2 robots collide with each other. The behavior of collision avoidance requires very precise and robust actions of the controller in high robot density, yet the neural network trained by DRL has poor generalization ability to achieve sufficiently safe collision avoidance. From \textbf{Table \uppercase\expandafter{\romannumeral1}}, SRL-ORCA substantially improves the success rate compared to DRL using the same network or Hybrid-RL ($\bar{\chi }^{srl-orca}=90.7\%,  \bar{\chi }^{drl-cd}= 58.0\%,  \bar{\chi }^{hybrid-rl}= 73.7\%$), where the dynamic collision ratio is reduced from $38.7\% $ to $2.7\% $. Since DRL and Hybrid-RL have similar trajectories, we only provide experimental pictures of DRL. This small number of collisions ($2.7\% $) is due to insufficient real-time computational power during the operation of SRL-ORCA. In our video, as the 8 robots cyclically birth from the starting point in Scenario-3, the number and relative positions of the robots that meet at the intersection change all the time. SRL-ORCA still guarantees a high success rate, which indicates that SRL-ORCA has stronger robustness than DRL. In summary, SRL-ORCA achieves cooperative behavior among robots compared to NH-ORCA, obtaining smoother trajectories and faster average speeds ($14.1\% $ faster than NH-ORCA). Moreover, SRL-ORCA absolutely overcomes the shortcomings of DRL in terms of collision avoidance (Task-3).

\begin{figure*}[htbp]
	\centering
	\vspace{-0.3cm}
\centering
\subfigure[]{
\begin{minipage}[b]{0.33\linewidth}
\centering
\includegraphics[width=1\textwidth]{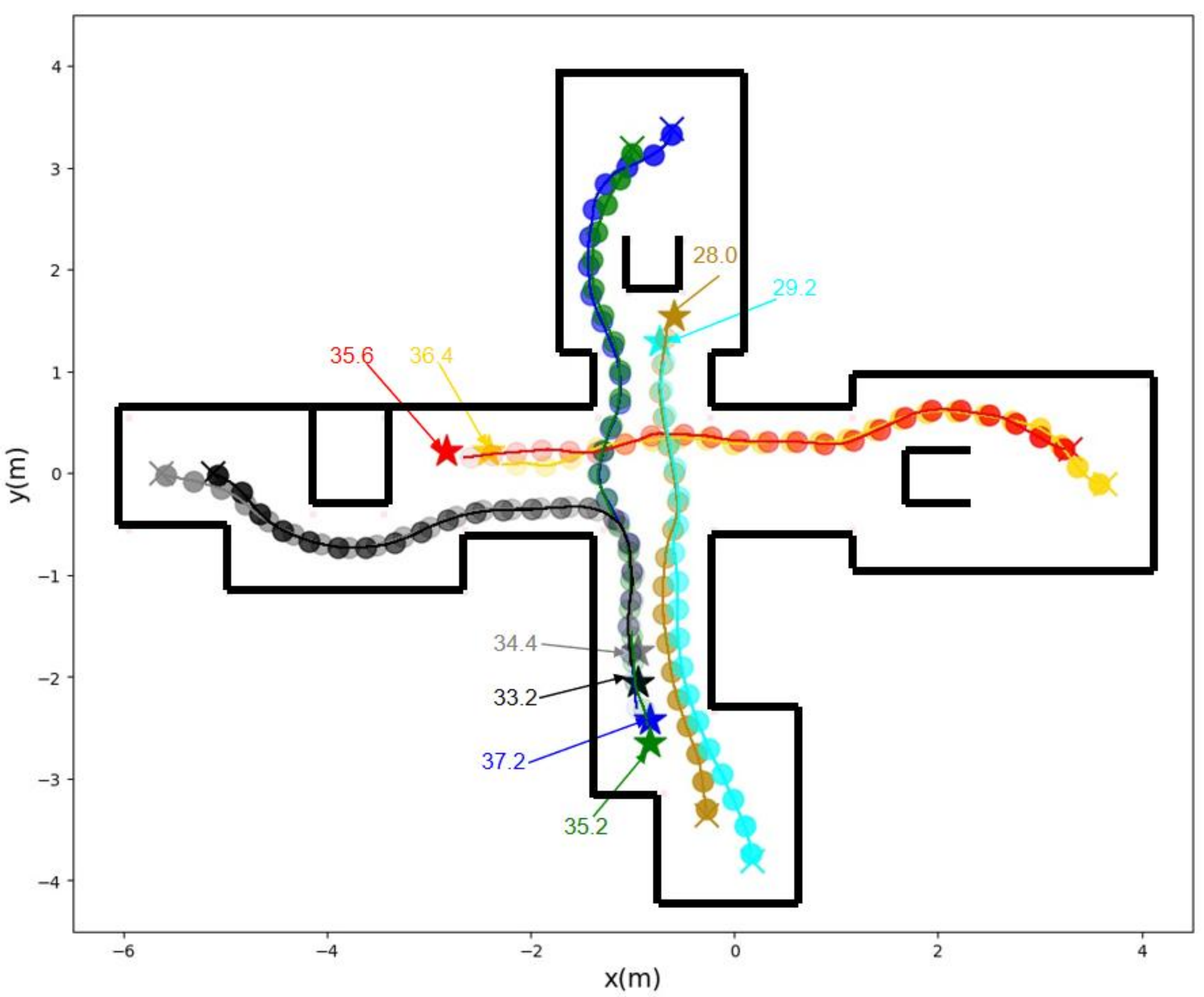}
\end{minipage}%
}%
\subfigure[]{
\begin{minipage}[b]{0.33\linewidth}
\centering
\includegraphics[width=1\textwidth]{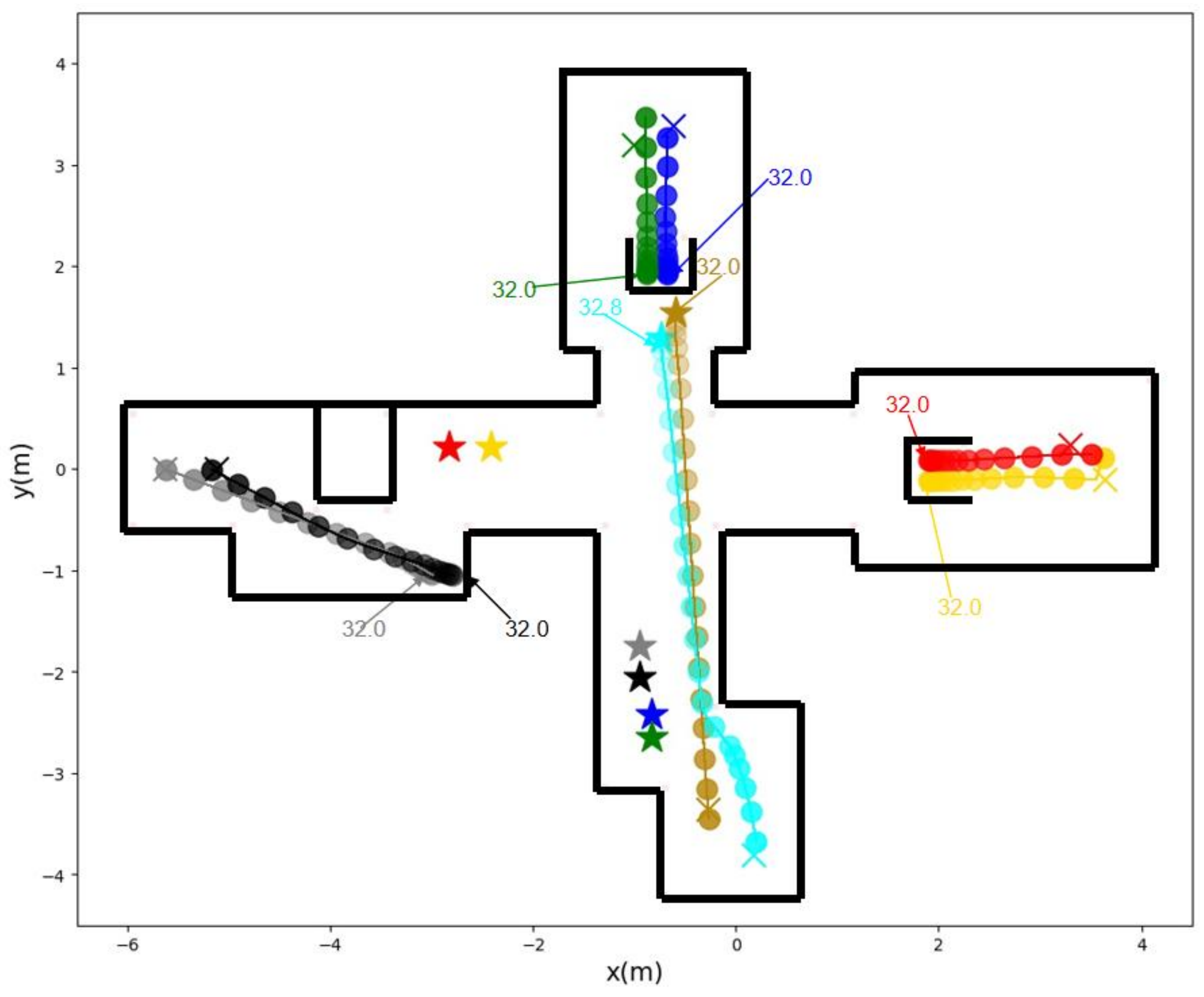}
\end{minipage}%
}%
\subfigure[]{
\begin{minipage}[b]{0.33\linewidth}
\centering
\includegraphics[width=1\textwidth]{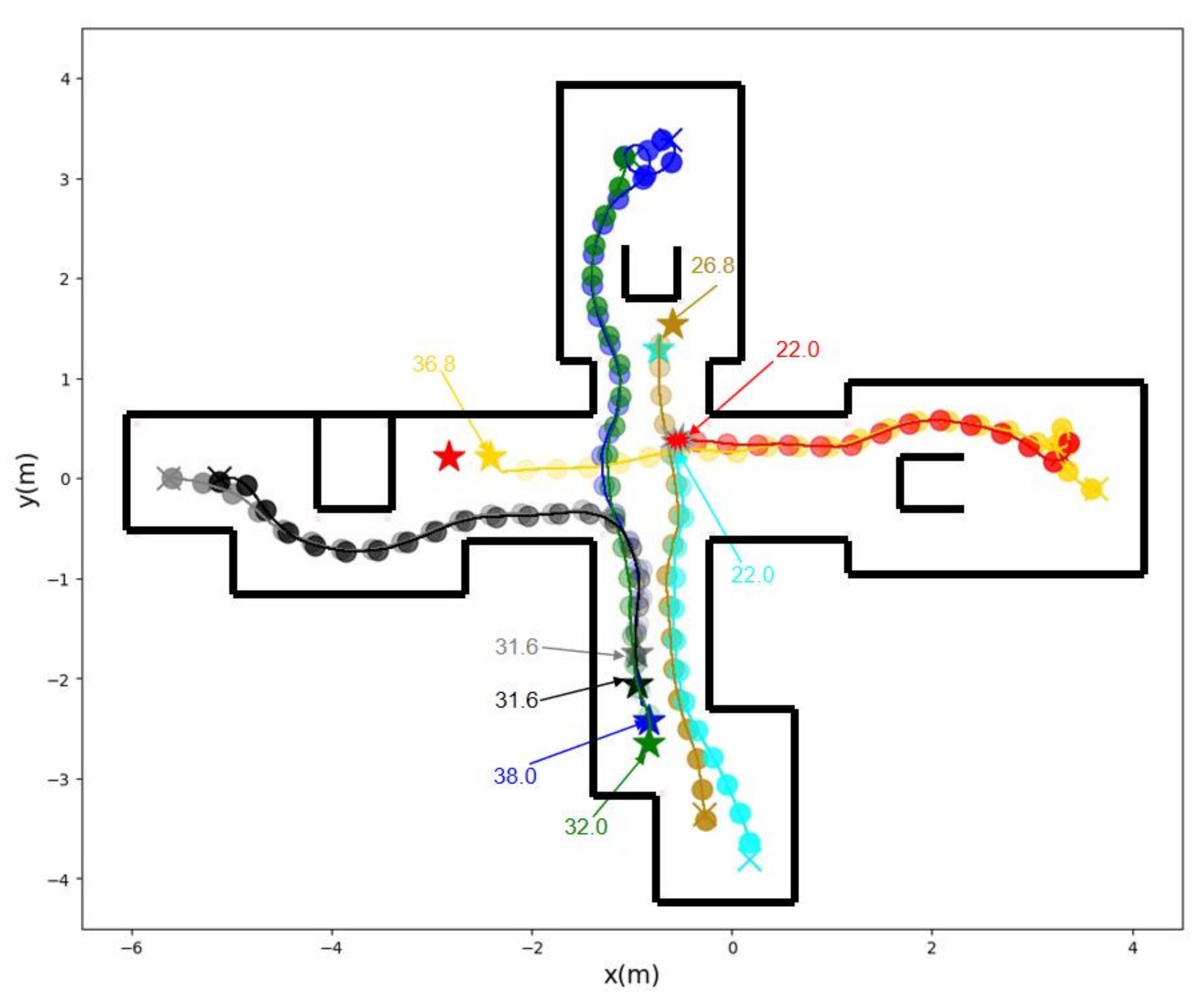}
\end{minipage}
}%

\centering

\caption{Scenario-4: 8 robots each start to cross the obstacle area ahead and then go through the intersection to reach the destination (6 straight ahead, 2 turn right). (a) shows that SRL-ORCA drives the robots to bypass the non-convex region and navigate smoothly. (b) shows that 6 robots are trapped in front of convex obstacles and only 2 robots with convex obstacles in front of them can move normally when using NH-ORCA. (c) shows that though the robots using DRL were able to avoid the non-convex region, 2 robots still collided at the intersection, and only 6 robots reached their destinations.}
	\label{fig13}
   	\vspace{-0.5cm}
\end{figure*}

\textbf{Scenario-4: }As in Fig.~7, this is a complex scenario mixed with dynamic and static non-convex obstacles. In Fig.~7(a), the policy $\pi _{\theta }^{srl-orca} $ obtained by learning in SRL-ORCA drives the robots to bypass the non-convex region and navigate smoothly ($\bar{\chi }^{srl-orca}=90.0\%$). At the intersection, the encounter robots also demonstrate full cooperation, driving on the right side and avoiding the meeting vehicles. As in Fig.~7(b), since NH-ORCA cannot deal with non-convex static obstacles, 6 robots freeze in front of non-convex obstacles, and only 2 agents with convex obstacles in front of them can move normally ($\bar{\chi }^{nh-orca}=25.0\%$). In Fig.~7(a)-(c) and \textbf{Table \uppercase\expandafter{\romannumeral1}}, SRL-ORCA achieves the highest success rate ($\bar{\chi }^{srl-orca}=90.0\%$) in this scenario than NH-ORCA, DRL, and Hybrid-RL ($\bar{\chi }^{drl-cd}=84.0\%, \bar{\chi }^{hybrid-rl}=82.7\%$). Results of Scenario-1,4 adequately demonstrate that the SRL-ORCA is more "intelligent" to handle complex non-convex scenes without maps. Fig.~7(c) shows that though the robots using DRL are able to avoid the non-convex region, two robots still collided at the intersection where multiple robots met, and only six robots reached their destinations. In summary, SRL-ORCA has better pathfinding performance (Task-1) in complex non-convex environments. Under Scenario-4, SRL-ORCA achieves the highest success rate $90.0\% $ compared to NH-ORCA, DRL, and Hybrid-RL.

{\color{green}}
In Scenario-1,2,3,4, SRL-ORCA accomplished the combined tasks of pathfinding (\textbf{Task-1}), obeying social norms (\textbf{Task-2}), and collision avoidance (\textbf{Task-3}) better in mapless navigation. Furthermore, the above scenarios all demonstrate that SRL-ORCA successfully follows specific social norms (\textbf{norm-1, 2, 3}). In dynamic complex environments, SRL-ORCA achieves the best navigation success rate ($90.7\% $ in Scenario-3, $90\% $ in Scenario-4). Compared to the DRL or Hybrid-RL algorithm, SRL-ORCA improves the success rate of obstacle avoidance; compared to the NH-ORCA algorithm, SRL-ORCA copes with non-convex obstacles better and improves the average speed ($14.1\%$ faster than NH-ORCA).

\section{CONCLUSION}

We propose a socially aware multi-agent mapless navigation algorithm with SRL-ORCA. SRL-ORCA realizes multi-agent safe reinforcement learning by using ORCA as an external knowledge to provide safety advice to the DRL. SRL-ORCA overcomes the shortcomings of obstacle avoidance in DRL and shows better performance in robot motion safety. It introduces traffic norms of human society to enhance social comfort and achieve cooperative avoidance. SRL-ORCA is able to cope with non-convex obstacles without falling into local minimal regions and achieves better trajectory quality in complex scenarios. Our future work will further investigate safe and efficient multi-agent cooperative navigation.

\bibliographystyle{IEEEtran}
\bibliography{rough}

\end{document}